%% file: acl_latex.tex
\pdfoutput=1
\documentclass[11pt]{article}

\usepackage[final]{acl}

\usepackage[a-1b]{pdfx} % For PDF/A-1b compliance

\usepackage{times}
\usepackage{latexsym}

\usepackage[T1]{fontenc}
\usepackage[utf8]{inputenc}

\usepackage{microtype}

\usepackage{inconsolata}

\usepackage{graphicx}

\usepackage{times}
\usepackage{listings}
\usepackage{multirow}
\usepackage{soul} % for strikethrough
\usepackage{subcaption}
\usepackage{graphics}
\usepackage{xspace}
\usepackage{amsmath}
\usepackage{amssymb}  % for checkmarks.
\usepackage{adjustbox}
\usepackage{makecell}
\usepackage{pifont}
\usepackage{tabularx}
\usepackage{xcolor}
\usepackage{booktabs}
\usepackage{array}

\newcommand{\ours}{{\textsc{CaT-Bench}}\xspace}
\newcommand{\gthree}{{\texttt{gpt-3.5-turbo}}\xspace}
\newcommand{\llm}{\textsc{LLM}\xspace}
\newcommand{\llms}{\textsc{LLM}s\xspace}
\newcommand{\gfturbo}{\texttt{gpt-4-turbo}\xspace}
\newcommand{\gfomni}{\texttt{gpt-4o}\xspace}
\newcommand{\gfmini}{\texttt{gpt-4o-mini}\xspace}
\newcommand{\oonepre}{\texttt{o1-preview}\xspace}
\newcommand{\llama}{\texttt{Llama3-8B}\xspace}
\newcommand{\gem}{\texttt{gemini-1.0-pro}\xspace}
\newcommand{\gemadv}{\texttt{gemini-1.5-pro}\xspace}
\newcommand{\flash}{\texttt{gemini-1.5-flash}\xspace}
\newcommand{\sonnet}{\texttt{claude-3.5-sonnet}\xspace}
\newcommand{\gfour}{\textsc{GPT-4}\xspace}
\newcommand{\tc}{\textsc{TC}\xspace}
\newcommand{\ndsc}{\textsc{OCC}\xspace}
\newcommand{\binarytask}{Step Order Prediction\xspace}
\newcommand{\humantask}{Step Order Explanation\xspace}
\newcommand{\dep}{\textsc{Dep}\xspace}
\newcommand{\ndep}{\textsc{NonDep}\xspace}
\newcommand{\ndeps}{\textsc{NonDep-S}\xspace}
\newcommand{\avg}{\textsc{Avg}\xspace}
\newcommand{\avgbin}{\textsc{AvgBin}\xspace}

\newcommand{\maj}{\textsc{MajVote}\xspace}
\newcommand{\modavg}{\textsc{ModAvg}\xspace}

\newcommand{\eat}[1]{}

\newcommand{\squishlist}{
  \begin{list}{$\bullet$}
    { \setlength{\itemsep}{0pt}      \setlength{\parsep}{3pt}
      \setlength{\topsep}{3pt}       \setlength{\partopsep}{0pt}
      \setlength{\leftmargin}{1.5em} \setlength{\labelwidth}{1em}
      \setlength{\labelsep}{0.5em} } }
\newcommand{\reallysquishlist}{
  \begin{list}{$\bullet$}
    { \setlength{\itemsep}{0pt}    \setlength{\parsep}{0pt}
      \setlength{\topsep}{0pt}     \setlength{\partopsep}{0pt}
      \setlength{\leftmargin}{0.2em} \setlength{\labelwidth}{0.2em}
      \setlength{\labelsep}{0.2em} } }

 \newcommand{\squishend}{
     \end{list} 
 }

\usepackage{todonotes}
\title{\ours: Benchmarking Language Model Understanding \\ of Causal and Temporal Dependencies in Plans}

\author{Yash Kumar Lal$^1$\thanks{Equal Contribution}, Vanya Cohen$^2$$^*$, Nathanael Chambers$^3$, \\ \textbf{Niranjan Balasubramanian$^1$, Raymond Mooney$^2$} \\
\\
  $^1$Stony Brook University,
  $^2$University of Texas, Austin\\
  $^3$US Naval Academy \\
  $^1$\texttt{\{ylal,niranjan\}@cs.stonybrook.edu},
  $^2$\texttt{\{vanya,mooney\}@utexas.edu},\\
  $^3$\texttt{nchamber@usna.edu}\\}

\begin{document}
\maketitle
\begin{abstract}
Understanding the abilities of \llms to reason about natural language plans, such as instructional text and recipes, is critical to reliably using them in decision-making systems. 
A fundamental aspect of plans is the temporal order in which their steps need to be executed, which reflects the underlying causal dependencies between them. 
We introduce \ours, a benchmark of Step Order Prediction questions, which test whether a step must necessarily occur before or after another in cooking recipe plans. 
We use this to evaluate how well frontier \llms understand causal and temporal dependencies. 
We find that SOTA \llms are underwhelming (best zero-shot is only $0.59$ in F1), and are biased towards predicting dependence more often, perhaps relying on temporal order of steps as a heuristic.
While prompting for explanations and using few-shot examples improve performance, the best F1 result is only $0.73$. Further, human evaluation of explanations along with answer correctness show that, on average, humans do not agree with model reasoning. 
Surprisingly, we also find that explaining \emph{after} answering leads to better performance than normal chain-of-thought prompting, and \llm answers are not consistent across questions about the same step pairs.
Overall, results show that \llms' ability to detect dependence between steps has significant room for improvement.
\end{abstract}

\section{Introduction}
% \vc{Consider delete: central to decision making and }
Planning is central to decision making and has been studied in various domains such as robotics and embodied environments \cite{planning-algo-book, jiang2019task}. 
% It requires the ability to reason about several aspects such as preconditions and effects, entity states and conditions, and temporal connectedness. 
% Even moderately complex plans require consideration of aspects like states, conditions, and temporal connectedness. 
To follow, revise, or customize a plan, one must be able to reason about the steps involved as well as their causes and effects \cite{plasma, lal-etal-2024-tailoring}.
Recent work on evaluating reasoning in plans focuses on classical problems such as Blocksworld \cite{slaney-blocksworld, valmeekam-etal-2023-planbench}, simulated environments like AlfWorld \cite{ALFWorld20}, or restricted language such as PDDL \cite{zhang2024proc2pddl}.
However, real-world natural language plans cannot be executed to test for correctness and reliability.
This paper describes a new question-driven evaluation to better study the detailed causal and temporal connections within such plans.
% \vc{detailed dependencies within such plans.?}
% Source: https://docs.google.com/drawings/d/1NJ5fqdyG9KSHo4cgkMye9MD79G41s38aE300V-Z2_vQ/edit?usp=sharing
\begin{figure}[!t]
    \centering
	\includegraphics[width=\columnwidth]{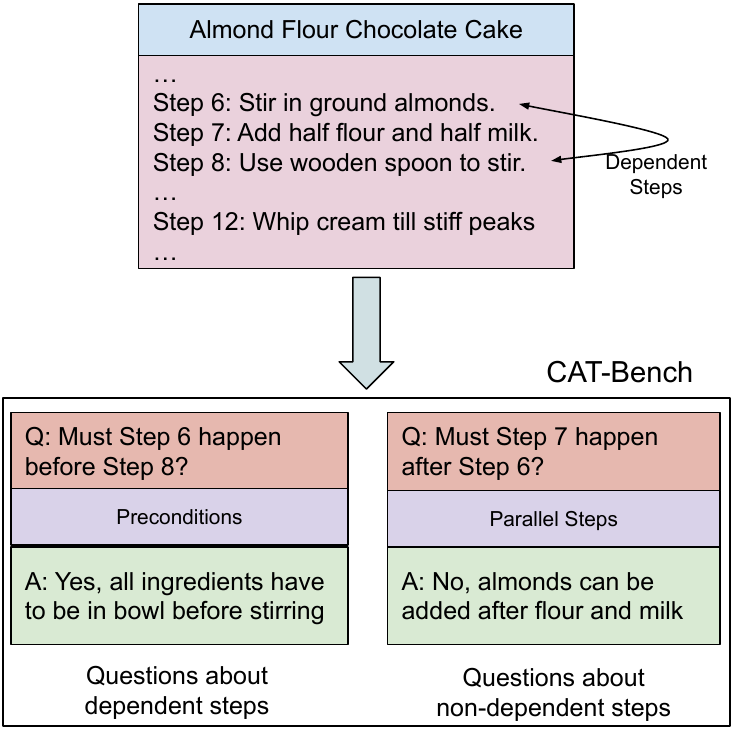}
	\caption{We use step-pair dependency annotations to create \ours, a question-driven evaluation framework for plan-based reasoning. Questions in this benchmark elicit reasoning about different causal relations such as preconditions, effects and step independence.
 }
	\label{fig:main-overview-fig}
\end{figure}

Given a plan, such as making a cake in \autoref{fig:main-overview-fig}, one must understand its various aspects to answer questions about it.
Answering if ground almonds should be added before stirring the mixture requires understanding that a \emph{precondition} for mixing evenly is that all ingredients should be added already.
But reasoning about whether flour should be added after almonds requires figuring out \emph{step independence} since the order of adding ingredients doesn't matter here.
% \vc{Suggestion: doesn't matter.}
Such causal aspects are encoded in temporal step dependencies of plans.
% To ensure reliability, we can use question-driven evaluation, for each aspect of a plan as shown in \autoref{fig:main-overview-fig}.\nb{This is a cursory reference to the figure. Say a bit more about the items in the figure.}
% For reliable planning, we need rigorous ways to evaluate each such aspect. 
% While large language models (\llms) demonstrate an impressive range of reasoning abilities, it is unclear how well they understand actions and their causes and effects in plans.
% \ykl{tried to address below comment}
% \nir{Can you build up the problem a bit before saying what you have done? Are there existing datasets for understanding plans? What are they missing? In what way you have been clever (if at all) in creating this resource? Discuss the notion of question-driven evaluation somewhere. Refer to the figure.}
% \mooney{I feel the focus is too much in this intro on "planning" which normally means constructing plans given an initial state and a goal.  I feel our work is more on plan understanding, which has its own motivations and applications. Can we at least balance this more to focus more on this? }
% \ykl{tried to address this comment}
% In this paper, we focus on the causal and temporal links between the plan steps.
We modify the Recipe Flow Graph Corpus \cite{yamakata-etal-2020-english}, containing recipes with substep procedure dependencies, to construct a new question-based dependency benchmark. 
% For each recipe, a directed acyclic graph (DAG) is constructed where nodes are recipe sentences and edges indicate dependencies between those steps, defining a partial ordering of the steps in the plan,
% We use 20\% of this data for evaluation.
\ours contains 4260 questions about causal dependencies spanning 57 unique plans.

\citet{west-etal-2024-the} show that \llms create expert-like outputs, but their generative capability does not necessarily indicate a correspondingly strong capability to understand underlying phenomena.
% \vc{may not reflect their underlying ability to understand?}
While \llms appear to generate good plans, it's unclear how well they understand important aspects of the steps themselves.
We thus use \ours to test whether \llms can identify step dependencies that reflect the causal and temporal structure of the plan.
We find that current \llms struggle to identify step dependencies, often performing close to random chance, raising more questions about their understanding of instructional text.
%In fact, their identification of independent steps is worse than random chance.

Using notions of consistency to evaluate their robustness, we also show that almost all out-of-the-box \llms are largely unreliable. Few-shot prompting with retrieved exemplars improves performance and consistency ($0.49 \rightarrow 0.68$ F1 for \gfomni).
Explanation-based generation offers another route to improve model performance and reliability on reasoning tasks \cite{camburu-etal-2018-esnli, rajani-etal-2019-explain, kumar-talukdar-2020-nile}. 
Prompting \llms to explain their decisions also improves performance on \ours ($0.49 \rightarrow 0.7$ F1). 
Despite these gains, there is still a large room for improvement in identifying step dependencies. 
When also considering the quality of the explanations, the average human ratings for satisfactory answers from SOTA \llm's is only $\sim$3 (out of 5).
%On average, even SOTA \llms' explanations that actually support the answer are scored $\sim$3 (out of 5) by humans.
Further, contrary to prior findings, using chain-of-thought prompting (CoT), i.e., reasoning before answering \cite{wei-etal-2022-chain}, performs worse than answering first and then explaining it, indicating inconsistencies in model reasoning.\footnote{\ours is available at \url{https://huggingface.co/datasets/vanyacohen/CaT-Bench} and the code is at \url{https://github.com/StonyBrookNLP/CaT-Bench}.}

In summary, this paper:
\squishlist
    \item Introduces \ours, a benchmark to evaluate the causal and temporal reasoning abilities of \llms over instructional plans.
    \item Demonstrates that current \llms cannot predict causal dependencies in plans well, and highlights what aspects are most difficult.
    \item Evaluates explanations for correctness and as a prompting mechanism to improve reasoning. %but their explanations are valid most of the time. \nc{put a validity number here?}
    \item Analyzes successes and failures of \llms, finding that generating a prediction followed by an explanation is significantly better than CoT.
\squishend

\section{Related Work}

% Starting point: https://docs.google.com/document/d/1QVXHK-as1mVCnIkAvRDRO4QeWXyZreenY4JQ-ycEe7Q/edit

Early work in text understanding argued for the importance of understanding plans and goals \cite{schank-abelson-1977-scripts}.
Generating plans \cite{aouladomar-saint-dizier-2005-towards} involves different types of understanding such as temporal reasoning and entity state tracking.
% With the advent of \llms, 
NaturalPlan \cite{zheng-etal-2024-naturalplan} present real-world tasks with natural language interaction, but are only limited to three tasks.
PlanBench \cite{valmeekam-etal-2023-planbench} showed that \llms were unable to generate executable and effective plans, but focused on simulated worlds with restrictive PDDL syntax.
\citet{lyu-etal-2021-goal} proposed the Goal-Oriented Script Construction task, where a model produces a sequence of steps (or a plan) to accomplish a given goal.
ChattyChef \cite{chattychef} uses the conversational setting to generate cooking instructions and iteratively refine its step ordering.
% For the task of counterfactual planning, 
CoPlan \cite{plasma} collects conditions associated with a revised list of steps for the task of plan revision to satisfy constraints. 
\citet{lal-etal-2024-tailoring} study the use of \llms for plan customization according to user requirements.
% \mooney{I don't know this work but I don't understand what the task is from this description}
\llms have been shown to generate plans well but it is unclear how well they truly understand all aspects of these plans.

Plan understanding tasks involve multiple aspects such as tracking entity states \cite{Bosselut2018SimulatingAD, henaff2017tracking}, linking actions \cite{pareti2014integrating, lin-etal-2020-recipe, donatelli-etal-2021-aligning}, next event prediction \cite{nguyen-etal-2017-sequence,zellers-etal-2019-hellaswag,zhang-etal-2020-analogous} and more.
OpenPI \cite{tandon-etal-2020-dataset, zhang-etal-2024-openpi2} enables entity tracking in how-to procedures.
ProPara \cite{dalvi-etal-2018-tracking} focuses on describing and understanding scientific processes.
XPAD \cite{dalvi-etal-2019-everything} extend ProPara by adding the new task of explaining actions by predicting their dependencies. 
\citet{zhang-etal-2020-reasoning} formalize several multiple-choice tasks related to step- and goal- relations in procedures.
\citet{kiddon-etal-2015-mise} explore predicting dependencies in cooking recipes and related tasks.
Similar work has been done on identifying dependencies in multimodal instructions with images and text \cite{pan-etal-2022-multimodal, wu-etal-2024-understanding}.
% \citet{zhou-etal-2022-show} construct an open-domain hierarchical knowledge-base (KB) of procedures based on wikiHow \cite{koupaee-wang-2018-wikihow}.
PizzaCommonsense \cite{diallo-etal-2024-pizzacommonsense} is a  dataset for learning commonsense about intermediate and implicit steps for cooking recipes, and contains explicit input/output pairs of each action with fine-grained annotations.
Choice-75 \cite{choice75} aims to study decision branching in plans by generating user scenarios and choices for various steps within the plan.
CREPE \cite{zhang-etal-2023-causal} measures how well \llms understand the comparative likelihood of two events occurring in a procedure.
There are a variety of datasets evaluating different aspects of plans, but there is a lack of one that clearly studies the prediction and explanation of temporal ordering constraints on the steps of an instructional plan.

\section{\ours}

% \nir{More examples would be helpful in the paper. You could add a table in this section with example questions and correct answers. Use the examples to illustrate what kinds of knowledge and reasoning models need to do to arrive at the answer.}
% describe data, creation process and attributes in detail here

% Taken from instruction unshuffling paper
% \nb{We are doing a bunch of interesting question types to get at the different aspects (e.g. preconditions, causes, effects) underlying dependencies. Let's build these up.}
% \ykl{can you join the meeting? need to discuss precision recall stuff}

Understanding plans requires reasoning about how different steps in a plan relate to each other.
In this work, we focus on the ability to recognize \emph{temporal dependencies} between steps i.e., deciding if a one step \emph{must} happen before another.
Typically, step $i$ must happen before a step $j$ if the effects (outcomes) of step $i$ satisfy one or more preconditions necessary for the proper execution of step $j$, or if the effects of step $j$ aggregate or modify the effects of step $i$ in service of accomplishing a (sub-)goal. 
For example, in the plan for baking shortcakes shown in \autoref{fig:dataset-examples}, step 10 which involves moving the (implicitly mentioned) baked cake to the wire rack for cooling, requires that the cake be baked first, which in turn requires the dough to be placed in the baking tray. %Similarly, step $j$ must happen after step $i$, if the effects of step $j$ aggregate or modify the effects of step $i$ in service of accomplishing the goal or a sub-goal in the plan. 
Thus, recognizing such dependencies requires the ability to infer many important logical connections such as preconditions, causes, sub-goals, and effects of the steps. 
This suggests that a simple test of whether a step must happen before another step (or after) can be an effective test of reasoning about the various logical dependencies between the steps in a plan. 

We build on this idea to create \ours, a new dataset of causal dependency questions defined over cooking recipes. 
Specifically, we make use of the Recipe Flow Graph Corpus \cite{yamakata-etal-2020-english} containing 300 English language cooking recipes annotated with substep procedure dependencies. 
For each recipe, this dataset provides a directed acyclic graph (DAG), in which the nodes are steps and directed edges indicate the temporal edge between those steps. 
If the nodes corresponding to two steps are not connected by a directed path, then they can be performed in any order (with respect to themselves) without changing the recipe result. 
In other words, two steps are temporally dependent if and only if there is a directed path from one to the other, and independent otherwise.

For all ordered pairs of steps $(i, j)$ in a plan, we create two binary (yes/no) questions: (i) \emph{Must $step_i$ happen before $step_j$?} (ii) \emph{Must $step_j$ happen after $step_i$?}. 
% \vc{For all ordered pairs of.}
These questions primarily test for for precondition relations (e.g. first question in \autoref{fig:dataset-examples}), and the ability to understand effects of steps and how they relate to sub-goals or overall goals of the plan (e.g second question in \autoref{fig:dataset-examples}).
% The former questions primarily test for precondition relations (e.g. first question in \autoref{fig:dataset-examples}). 
% The latter primarily test for ability to understand effects of steps and how they relate to sub-goals or overall goals of the plan (e.g second question in \autoref{fig:dataset-examples}). \mooney{Again, I just don't agree with this distinction, I would leave the last two sentences out as confusing and unnecessary.}
We pool all such questions from dependent pairs of steps (i.e, the steps where there is a directed path from one step's node to the other in the recipe DAG) 
% whose nodes share an edge in the recipe DAG \mooney{Again, I don't like this "share an edge" stuff, its about a directed path. }) 
into \dep, and the rest into \ndep.\footnote{Note that the answers to all the questions in the \dep set are `yes', and the answers to \ndep questions are `no'.}

%and answering them requires understanding preconditions, causes and effects. Answering \dep question in \autoref{fig:dataset-examples} requires understanding that placing dough on a tray is a precondition to remove a baked cake onto a wire rack.

%We do the same for any step pair $(step_i, step_j)$ that is explicitly annotated to not have a dependency between them, associate it with the corresponding plan, and refer to these questions as \ndep, which often requires understanding distinct subgoals.
% \nb{Why are we creating this swapped version? Give some motivation.}
% \nc{`For another twist on this task, and since...'} 
%Understanding step independence may involve identifying distinct subplans within a plan such as in \autoref{fig:dataset-examples}, figuring out that two steps can be performed in parallel or in any sequential order, or realizing that some steps can be optional.
% We present examples of each type of question in \autoref{fig:dataset-examples}.

\begin{figure}[!t]
\small
\centerline{
\fbox{
    \parbox{0.95\columnwidth}{
	{\bf Goal:} lemon zested strawberry shortcakes \\ 
    {\bf Steps:}\\
    ...\\
    6. Divide dough in half.\\
    7. Add sugar; beat until stiff peaks form.\\
    8. Place 5cm apart on an ungreased baking tray.\\
    9. Bake at 200 C / Gas 6 for 8-10 minutes.\\
    10. Remove to a wire rack; cool for 15 minutes.\\
    11. In bowl, combine butter and lemon zest; set aside.\\
    12. In mixing bowl, beat cream until it begins to thicken.\\
    13. Gently pat or roll each half into a 1.75cm thick circle.\\
    14. To assemble, split shortcakes in half.\\
    ...
    \squishlist
    \item[{\bf \dep Q:}] Must Step 8 happen before Step 10?
    \item[{\bf \dep A:}] Yes, removing a cake for cooling needs dough to be placed on a baking tray first.
    \item[{\bf \ndep Q:}] Must Step 12 happen after Step 7?
    \item[{\bf \ndep A:}] No, adding sugar is a part of making dough but beating the cream makes the filling.
    \squishend
    }
}}
\caption{Examples of different types of questions in a plan from \ours. To correctly answer these questions, one must understand preconditions and effects (to answer \dep), some steps need not be performed in any particular order and that plans can contain subplans within them (to answer \ndep).}
\label{fig:dataset-examples}
\end{figure}

\input{tables/prf}

In total, \ours contains 2,840 questions about causal dependencies of steps for 57 unique plans. 
We undersample the non-dependent questions to ensure that \ndep and \dep are of the same size (i.e., 1,420 questions each). 
%It contains 1,420 questions of each type (\dep and \ndep), is class-balanced.
Half of \ours tests the ``before" temporal relation and the other half tests the ``after" relation.
It is, thus, balanced in terms of both question types and temporal relation type.
We also annotate the questions based on the distance between the pairs of steps. 
Two steps are deemed close if they are within 3 steps of each other, $(j-i) \le 3$, there are 1,256 questions about \emph{close} steps and 1,584 about \emph{distant} steps.

\ours enables two tasks.
\binarytask~ elicits binary judgments about dependencies between pairs of steps in a plan, and performance on this task can be evaluated automatically.
\humantask~ requires models to provide explanations for their judgments about step dependencies.
This involves understanding causal relationships and expressing relevant knowledge about actions in the steps being asked about.
Since this is a free-form generation task, these explanations require human evaluation.
Note that \ours does not contain gold, human-written explanations, and we advocate for reference-free human evaluation since there can be multiple valid explanations.

\subsection{Automatic Metrics}
\label{subsec:auto_metrics}

We evaluate model performance on \binarytask~ on standard metrics of precision, recall and F1 score.
We measure robustness of these models using two metrics of consistency.
Models must provide consistent predictions when asked before/after questions about the same step pair, i.e., if a model judges that $step_i$ happens before $step_j$, it must also judge that $step_j$ happens after $step_i$. 
% \nb{This is slightly different in semantics from the ``must'' framing. When we ask ``Must B happen after A'', it seems a stricter framing than judging that ``B happens after A''}. \yash{clarified in meeting and uncommenting accordingly}
We define this metric as Temporal Consistency (\tc).

\subsection{Human Evaluation}
% \nate{\humantask has never been defined. This comes out of nowhere. Do we need a "this is the task" subsection? Section 3 defines CAT-Plan and the step order task, but nothing about explanations.}
For open-ended text generation tasks such as \humantask, the absence of an automatic metric that correlates well with human judgments is a major challenge \cite{chen-etal-2019-evaluating, ma-etal-2019-results, caglayan-etal-2020-curious,howcroft-etal-2020-twenty}.
So, we utilize human evaluation with a standardized interface to compare different models. 
We aim to measure whether a model output is a valid explanation for the given question.  
We present answers from different models and ask crowd-workers on Amazon Mechanical Turk to assess their correctness.
Workers are asked to rate the validity of each answer on a 5-point Likert scale \cite{likert1932technique} (1 to 5)\footnote{Integer scores correspond to the labels: strongly disagree, disagree, neutral, agree, strongly agree.}.
For each plan, question, and model answer, we ask 3 distinct annotators to provide judgments. 
% \vc{Suggest delete: distinct}
% \yash{indicates k-way HIT}
An explanation is considered invalid if it does not give a plausible reason that is also relevant to the question.
We provide more details in \autoref{appsec:human_eval}.
% A sample HIT can be seen in \autoref{fig:crowdsourcing-interface}.

% \input{tables/d_nd_nds}

\section{Benchmarking Models on \ours}

We benchmark the performance of a variety of models on \ours.

\subsection{Models}

We evaluate \gfturbo, \gthree, \gfomni, \sonnet, \gem, \gemadv, \flash, \gfmini and \llama.
These represent a diverse set from different model families and sizes. We evaluate them primarily in zero-shot prompting modes. We consider two settings: (i) generating only an answer (A), and (ii) generating an explanation along with the answer (A + E). 
The latter represents answering the question and then generating an explanation for it.
% \vc{I think we need to make clear that we also evaluate generic CoT and contrast this to A+E}
% \nc{add a sentence that is crystal clear if this is A-then-E or E-then-A}
We also analyze few-shot results for the answer-only (A) setting, and evaluate generic CoT (E + A) prompting. 
More details about each model can be found in \autoref{appsec:benchmark_models} and the prompts used in \autoref{appsec:prompts}.

\subsection{How Good Are Model Predictions?}
% \vc{Evaluating Model Predictions}

\autoref{tab:auto_eval} presents the performance of all the models in different settings on \binarytask.
We present per-class (\dep and \ndep) precision, recall and F1 score as well as macro average metrics on the class balanced \ours. We make three main observations.

%To start, we prompt each model to just produce an answer (A) in a zero-shot setting.

\noindent\paragraph{Models struggle at predicting step order.} In the zero-shot answer-only setting (A), \sonnet records the highest F1 score overall of 0.60. 
\gthree and \gfturbo are close behind with 0.59 and 0.58 respectively.
Surprisingly \gfomni, the most recent frontier model, fares significantly worse at 0.49 F1. 
It's smaller version, \gfmini, also performs similarly.
All three Gemini models (\gthree, \gem, and \flash) also only manage an F1 of around 0.55. 
\llama also fares poorly with an F1 of 0.49. 
Most models are comparable or barely better than a random baseline F1 of 0.5 on this balanced dataset showing that they are not able to directly answer the dependence question.

%\gthree, \gem, and \flash are the most consistent models with balanced macro average.
%Surprisingly, \gfomni performance is similar to \llama, even though prior work has shown that it should be significantly better due to its size and factors such as pretraining data.\gfour models are equivalent in terms of precision, while \gthree and \gfturbo achieve highest recall.Overall, we note that there is significant for improvement for all models on \binarytask, indicating that, in their vanilla form, they do not yet understand the underlying causal dependencies in a plan.

\noindent\paragraph{Generating explanations improves performance.}
Results for adding explanations to answers is shown in the (A + E) rows in \autoref{tab:auto_eval}.
Seven of the nine models, \gthree and \flash being the exceptions, have higher performance when also generating explanations. The biggest improvement in F1 is seen in \gfomni (+0.21). With explanations, the best result is the 0.73 F1 when using \gemadv. While this is substantially better than a random baseline, there is still significant room for improvement.

\noindent\paragraph{Models are biased towards predicting dependence.} 
%On \dep questions, \gemadv, \gfomni and \gfturbo achieve similar F1 scores when predicting and explaining.
%Looking closer at the answer-only (A) setting, we also find that \gfomni has very high recall.
Most models exhibit a higher recall for the \dep set and significantly lower recall for the \ndep set. 
This is particularly true for the answer only setting ((A) rows), the exceptions being \gthree and \gem. 
Coupled with the substantially lower precision values on the \dep set, this suggests that most models exhibit a bias towards predicting dependence between any given pair of steps. 
We hypothesize that they use temporal order of steps as a heuristic i.e, if a step appears before another step it is more likely to be dependent than not, and thus becoming biased towards predicting dependencies.
%Surprisingly, \gem and \gthree performance on \dep questions is worse than on \ndep questions. Particularly, both models show low recall, maybe not using the temporal order heuristic.

%Bias towards predict dependencies implies detecting non-dependent step pairs, since they are complementary relationships.
As noted earlier, using explanations improves the overall performance, translating to more balanced precision/recall values on \dep than when predicting answers alone. 
% \vc{Suggest delete: As we noted above}
Since the bias towards \dep necessarily means bias against \ndep, reduction in bias towards \dep also translates to a more balanced performance on both \dep and \ndep sets.
% \mooney{This last sentence seems obvious and could be deleted} 
However, even with explanations, the bias towards predicting dependence still remains to some extent for all models. 
Explanations improve \gfomni performance the most (+0.36) on \ndep questions. 
They do not help smaller models (\llama) identify dependencies better.

%This is in part because its answer-only setting (A) achieves very low recall, indicating that it is biased toward one label.

%\gem, \gthree and \flash performance decreases notably when also generating explanations.
%We find that \gemadv achieves very high precision, with the next model, \gfomni, significantly far behind. 
%Explanations do not help smaller models (here, \llama) identify dependencies better.

\subsection{How Good Are Model Explanations?}
% \vc{Evaluating Model Explanations}

On a random subset of 480 questions (240 \dep and 240 \ndep), we conduct a crowdsourced human evaluation of the explanations generated by \gfomni, \gfturbo, \gemadv and \llama, the three best \llms for \binarytask and an open-source model.
Annotators rate how much they agree (1 to 5) with the fact that the answer contains all the relevant details to address what the question requires.

% \nb{What is the statement with which the annotators agree or disagree when rating? Mention the statement here.}
%Annotators score model explanations on a Likert scale \cite{likert1932technique} of 1 to 5\footnote{Integer scores correspond to the labels: strongly disagree, disagree, neutral, agree, strongly agree. (Ray: this is all said aerlier}.
For each explanation, we compute the mean Likert rating from three distinct annotators. 
First, we report \avg, the overall average of these mean ratings across all 480 instances. 
To account for cases where the answer is incorrect, 
% \nb{This is still not correct. You are not doing this to correct for inconsistency between decisions and explanations. Rather, you are discounting explanations when the answer is wrong. Because on the whole, when the answer is wrong it doesnt matter what the explanation is. Ray's point later states that even when its answer is wrong, model's explanations can be convincing. I'd rephrase this to instead say, ``To account for cases where the answer is incorrect, we also devise''}, 
we also devise a new metric \modavg that accounts for cases where the step order prediction is incorrect. 
% \nb{Saying we are checking faithfulness seems a bit misleading and confusing. Are the annotators judging the explanation without knowing the model's decision? Now there are two aspects to this: 1) Suppose the models decision is incorrect e.g. say it incorrectly predicts dependence. Now the explanation also supports dependence and is thus faithful to its decision, which means the explanation supports an incorrect decision. So it is not a question of faithfulness that you are measuring. You are simply ignoring explanations where the models prediction is incorrect, which is a fine thing to do but it is not measuring faithfulness. 2) What if the annotators like this faithful explanation of an incorrect decision, however, then what do we do? I think our decision here is to ignore the ratings of explanations of an incorrect decision. In fact it seems to me that we should not show the explanation ratings for cases where the answer is incorrect. It just raises questions about how to rate an explanation when the answer is incorrect. How many rows would we be left with if we had only evaluated explanations where the answers were correct. In the table, lets instead show the stddev as another column instead of AVG.} 
% \mooney{Again, I also find the term "faithful" here confusing. I think it is meaningful to say that an explanation for a wrong answer is "valid" and we already asked the judges for this, so we should include both and comment on how models are actually not that bad at explaining their wrong answers}
To calculate \modavg, we modify \avg by zeroing out human judgments for explanations where the corresponding prediction is incorrect.

We use weighted Fleiss Kappa to calculate inter-annotator agreement.
The weighted agreement score on a 5 point scale was 0.76, indicating high agreement between annotators.
Details about the calculation can be found in Appendix \ref{subsec:fleiss}.

\input{tables/humaneval}

% \vc{Model size improves AVG score. Thus, human annotators are able to differentiate explanation quality across model performance levels. Therefore our task }

\autoref{tab:all_human} presents the quality of model generated explanations as judged by human annotators.
As expected, larger models are clearly better than the much smaller \llama on all metrics.
There is very little difference between the frontier models, \gfomni, \gfturbo and \gemadv.
\avg performance indicates that there is significant room for improvement in the quality of model explanations.
On \modavg, we see that even the best model performance is below 3 (`neither agree nor disagree' with a model's explanation).
By this metric, \gemadv explanations are worse than \gfour even though it generates more correct answers.
The difference between \avg and \modavg indicates models are capable of generating convincing explanations for their wrong answers.
They produce explanations which justify the opposite of their answer a significant number of times.
In fact, \llama does so almost half the time.
These results show that models have a lot of room for improvement in their ability and reliability to reason about step dependencies in plans.
%There is a lot of room for improvement in their reasoning about step dependencies in all models.

\section{Analysis}

To better understand the strengths and weaknesses of these models, we analyze their performance on \ours organized by different characteristics of the questions and model prompts.

\subsection{Robustness of Models}

\autoref{tab:robustness} presents two measures of consistency to quantify the robustness of the models, similar to \cite{verma-etal-2023-evaluating, elazar-etal-2021-measuring}.

\begin{figure}[!t]
\small
\centerline{
\fbox{
    \parbox{0.95\columnwidth}{
	{\bf Goal:} lemon zested strawberry shortcakes \\
    \hrule
    \vspace{0.1cm}
    {\bf Steps:}\\
    ...\\
    6. Divide dough in half.\\
    7. Add sugar; beat until stiff peaks form.\\
    8. Place 5cm apart on an ungreased baking tray.\\
    ...\\
    12. In mixing bowl, beat cream until it begins to thicken.\\
    13. Gently pat or roll each half into a 1.75cm thick circle.\\
    ...
    \squishlist
    \item[{\bf \ndep Q:}] Must Step 12 happen after Step 7?
    \item[{\bf \ndep A:}] No, adding sugar is a part of making dough but beating the cream makes the filling.
    \squishend
    \hrule
    \vspace{0.2cm}
    {\bf Steps:}\\
    ...\\
    6. Divide dough in half.\\
    7. In mixing bowl, beat cream until it begins to thicken.\\
    8. Place 5cm apart on an ungreased baking tray.\\
    ...\\
    12. Add sugar; beat until stiff peaks form.\\
    13. Gently pat or roll each half into a 1.75cm thick circle.\\
    ...
    \squishlist
    \item[{\bf \ndeps Q:}] Must Step 12 happen after Step 7?
    \item[{\bf \ndeps A:}] No, making the filling with cream can be done in parallel to adding sugar in the dough.
    \squishend
    }
}}
\caption{Since two steps that are not dependent on each other can be performed in any order, we swap their order in the plan and ask binary questions about them similar to \ndep. Note that, while the plan itself is altered, the question remains the same.}
\label{fig:nds-dataset-examples}
\end{figure}

\input{tables/consistency}

\noindent\paragraph{Temporal Consistency} 
For a pair of steps $(step_i, step_j)$, the answer to must $step_i$ happen before $step_j$ should be the same as the answer to must $step_j$ happen after $step_i$ regardless of question type.
As described in \autoref{subsec:auto_metrics}, we measure this notion of consistency through \tc.
% \mooney{You already explained this earlier, you could leave these intro sentences out}
% \nc{I think the repeat is helpful.}
We make two main observations: 
(i) Even the most consistent models \gfomni, \gemadv and \sonnet change their answers to the before and after versions of questions in 20+\% of the cases. The rest are far more inconsistent with \flash changing its answers for more than 55\% of the questions; (ii) Surprisingly, adding explanations reduces answer consistency for most models, with \gemadv (+24\%), \gfturbo (+14\%) and \sonnet (+31\%) being the only exceptions showing improved consistency upon generating explanations.

%I suggest discussing TC first before OCC. 
% \input{figs/before_after_fig}

\noindent\paragraph{Order Contrastive Consistency} Since step pairs without dependencies can be performed in any order, we introduce a twist on \binarytask in which the step pairs in \ndep are switched in the plan itself.
% \footnote{It's possible that artificially changing the `normal' flow of steps alters dependency-related knowledge.\mooney{I dont understand this footnote or see what it is adding}}
For each modified plan, we create similar binary questions to \ndep and refer to them as \ndeps.
This helps test whether a model uses the step order as a heuristic to answer the question.
% \nc{Why? Need a why ... how about `This helps test whether plan order influences the model's output.'}
We show an example in \autoref{fig:nds-dataset-examples}.

% We find that most models perform better on \ndeps than they do not \ndep.
% Due to the step swap operation, it is possible that the temporal order heuristic does not apply.
% Looking closer at the recall in \autoref{tab:nds_prf}, we find that it is higher on \ndeps than it is on \ndep, confirming that models do indeed pick up on temporal order and mistake it for dependencies.

% When asked about step pairs that are not dependent on each other, models must provide the same answer regardless of the ordering of the steps in the plan.
The answer to dependency questions about an independent pair of steps should stay the same regardless of the order in which the steps are presented in the plan.
Order Contrastive Consistency (\ndsc) measures the fraction of times models provide consistent answers to the same question across \ndep and \ndeps.
We observe a similar overall inconsistency on \ndsc as with \tc, even from the best models.
% \vc{Would use a word other than ``trend'' unless we're talking about an independent + dependent variable.}
% \nc{What is the general conclusion? How about `We observed a similar overall trend of inconsistency like with TC, even from the best models.' }
For most models, generating explanations hurt consistency.
% \vc{generating explanations}
Surprisingly, \llama is the most robust according to \ndsc even though its task performance is lowest.
In contrast, \gemadv, which has the highest task performance, is the least robust as per this metric.

% We refer to this metric as Order Contrastive Consistency (\ndsc).
% \nc{This confused me. Do the \ndsc numbers only evaluate the NONDEP subset? Do the \tc only evaluate DEP? Can we have a number that includes everything together with consistency? You can say "No-No" (NONDEP) or "Yes-No" or "No-Yes" (DEP) but never "Yes-Yes".}
% Finally, all the models are more accurate on \ndeps questions as compared to \ndep with highest achieved accuracy being $\sim$84\%, even though the only difference is that the relevant steps are swapped in the presented plan.
% They are also more robust to the temporal relation in the question.
% This highlights another inconsistency of \llms that we measure through the \ndsc metric.
% \nate{It's possible this is due to the disruption of the `normal' recipe flow by moving the steps artificially.}

\subsection{Chain-of-Thought Struggles}
% \vc{Chain-of-Thought Limitations}
% \vc{Chain-of-Thought Under-performance}

In the experiments thus far, we have asked models to generate explanations for their answers. 
This can be seen as a answer-then-explain (A + E) approach. 
In contrast, the standard Chain-of-Thought (CoT) prompting strategy \cite{wei-etal-2022-chain} can be seen as an explain-then-answer approach (E + A), where we ask the model to generate reasoning or explanation that leads to its answer. 
In practice, this step-by-step reasoning can be seen as allowing the use of intermediate decoding tokens (like a scratchpad) in service to coming up with a possibly more accurate final answer for many tasks. \autoref{tab:binary_explain_vs_cot} compares performance of CoT prompting\footnote{We tried multiple CoT prompts, all with $temp=0$, but it had little effect on performance.} (E + A) to first predicting the answer and then explaining it (A + E) and simply providing an answer (A), all in the zero-shot setting when using \gfomni.\footnote{We use \gfomni instead of \gemadv due to rate limits on the latter.}

\input{tables/cot}
% \nate{Since this is sort of a negative result, would it help to say something like "We experimented with a number of prompts to request the explanation and prediction, and it had little effect on performance (or we chose the best performing)" ... if either is true of course.}

While chain-of-thought (E + A) results in an improvement over the answer-only setting (A), its performance is far below its counterpart (A + E).
This contradicts the expectation that it is better to use CoT for intermediate reasoning rather than answering and then generating explanations.
However, we note that (E + A) does lead to the highest temporal robustness amongst all approaches. 
% \nir{This is an interesting result. Can we make a bigger deal about this? What do the explanations look like in either case? Are they different in some systematic way? Why does the temporal consistency improve? }
% \yash{addressed}

Looking closer, \autoref{tab:d_cot} shows the performance of these methods on \dep questions.
CoT (E + A) has a higher bias towards predicting dependencies compared to both (A) and (A + E).

\input{tables/d_cot}

Further manual analysis of the explanations shows that (E + A) generates more vacuous explanations (e.g. just stating that a pair of steps are independent without digging deeper into what makes them independent).
Additionally, (E + A) makes errors by misunderstanding entity states for a step.
% \nb{This is circular. You were trying to explain why CoT doesnt help. Saying that it generates wrong explanations because it answers are incorrect is not an explanation of why it is incorrect.}
Unsurprisingly, both settings suffer from some hallucinations, sometimes missing the content of the steps being asked about.
We present an example of such hallucinations in \autoref{fig:hallucinations}.
% \nb{How about the following as a conclusion to this paragraph?}
These results are further indicators of brittleness and inconsistencies in models' reasoning about step dependencies. 
% Most models are able to accurately answer questions about effects (\textit{after}) more than ones about causes and preconditions (\textit{before}), as seen in \autoref{fig:before_after_binary}.
% Only \gfomni and \gemadv are different.
% We hypothesize that this is because effects in recipes can be more immediate and hence, would be easier to understand. \nir{Can you find examples of this to illustrate this point?}

\begin{figure}[!t]
\small
\centerline{
\fbox{
    \parbox{0.95\columnwidth}{
	{\bf Goal:} lightly spiced tomato and red lentil soup \\ 
    {\bf Steps:}\\
    1. Heat the oil in a saucepan.\\
    2. Add onion and cook over low heat for 7-8 mins...\\
    3. Stir in celery and carrot and cook for 3 mins...\\
    ...\\
    9. Blend the soup using hand-held blender.\\
    10. Check the consistency.\\
    11. Season to taste.\\
    ...
    \squishlist
    \item[{\bf Question:}] Explain why or why not Step 10 must happen after Step 3.
    \item[{\bf Model Answer:}] Step 10 does not need to happen after Step 3 because the eggplant needs to be cooked before blending.
    \item[{\bf Human Score:}] 1.0
    \squishend
    }
}}
\caption{Example of hallucinations produced by \gfour in the (E + A) setting.}
\label{fig:hallucinations}
\end{figure}

We also include zero-shot results with \oonepre, which was released just before the time of publication.
\oonepre uses search over chain of thought explanations as part of its inference process.\footnote{\url{https://openai.com/index/introducing-openai-o1-preview/}}
On \ours, this model achieves state-of-the-art performance scoring 0.80 F1.
This is better than both zero- and few-shot performance of any other model, and even achieves the best \tc at 85\%.
However, even this powerful model shows a bias towards predicting dependence (F1 of 0.83) between steps more than their non-dependence (F1 of 0.76).
Due to rate limits and prohibitive costs (\$32 for each \binarytask experiment), we were unable to investigate \oonepre further.

% \autoref{fig:before_after_binary_expl} shows that most models understand both temporal relations better when first predicting the dependency and then explaining their prediction (A + E).
% Different from (A), we note that \gemadv performs better on \textit{after} questions.
% Surprisingly, \gem and \gthree go down in performance on both aspects.

% \subsection{Do Known Prompting Techniques Help?}
\subsection{Effect of Improved Prompting Techniques}
% \vc{Suggest: Improved Prompting Techniques}

We also experiment with self-consistency \cite{mitchell-etal-2022-enhancing} and few-shot prompting (or in-context learning) \cite{brown-etal-2020-fewshot, incontext} on \gfomni (A)\footnote{Due to the lack of gold explanations, we are unable to run these variations for (A + E) or (E + A).} for the \binarytask.

For self-consistency, we use k=$\{3,5\}$ and temperature=$\{0.6, 0.8\}$ to sample binary predictions and take the majority of the predicted labels as the model's final answer.
\autoref{tab:prompt_variants} shows the results for one setting. 
Contrary to previous findings \cite{mitchell-etal-2022-enhancing}, self-consistency does not provide any improvement over the vanilla zero-shot model performance.
We report performance with other parameters in \autoref{tab:prompt_variants_full}.

\input{tables/prompting}

We use in-context learning \cite{incontext} with examples selected from the balanced training set using the BM25\footnote{\url{https://github.com/dorianbrown/rank_bm25}} \cite{bm25} algorithm.
We use k=5 exemplars and dynamically retrieve exemplars from a held-out set that are closest to the test instance.
As expected, few-shot prompting improves binary prediction performance a lot (+0.19).
In fact, few-shot performance is almost as good as predicting then explaining with \gfomni.
% The increase in accuracy on both \ndep and \ndeps questions is $\sim$20\% each.
% Even on \dep questions, where zero-shot \gfomni already achieves $\sim$92\% accuracy, we find that in-context learning provides a boost of $\sim$5\%.
% \nir{Make a note of order contrastive consistency dropping/not improving.}
% \nate{But OCC doesn't improve. NONDEP does. Does this mean it's inconsistent on the NONDEPs, just starts tagging more of them? Seems like there is something to highlight here about the challenge.}

\subsection{Error Analysis}

\begin{figure}[!t]
\small
\centerline{
\fbox{
    \parbox{0.95\columnwidth}{
	{\bf Goal:} vanilla rye cheesecake \\ 
    {\bf Steps:}\\
    1. Preheat the oven to 150 degrees C\\
    2. Place the biscuits into a plastic bag and crush.\\
    ...\\
    10. Place baking dish onto bottom rack of preheated oven and fill halfway with boiling water.\\
    ...
    \squishlist
    \item[{\bf Question:}] Explain why or why not Step 10 must happen after Step 2.
    \item[{\bf Model Answer:}] Step 10 must happen after Step 2 because the baking dish with boiling water needs to be placed in the oven just before baking the cheesecake to create a moist environment, which is necessary after the biscuit base has been prepared and the filling has been added.
    \item[{\bf Human Score:}] 5.0
    \squishend
    
    \hrule
    \vspace{0.2cm}
    {\bf Goal:} lemon zested strawberry shortcakes \\ 
    {\bf Steps:}\\
    1. In bowl, combine strawberries and sugar...\\
    2. In bowl, combine flour, sugar, baking powder and salt.\\
    3. Rub butter into flour until forming coarse crumbs.\\
    ...\\
    9. Bake at 200 C / Gas 6 for 8-10 minutes.\\
    10. Remove to a wire rack; cool for 15 minutes.\\
    11. In bowl, combine butter and lemon zest; set aside.\\
    ...
    \squishlist
    \item[{\bf Question:}] Explain why or why not Step 10 must happen after Step 2.
    \item[{\bf Model Answer:}] Step 10 involves cooling the baked shortcakes, which are prepared and baked in steps 5 through 9, and does not depend on the flour mixture preparation in step 2.
    \item[{\bf Human Score:}] 1.0
    \squishend
    }
}}
\caption{Examples of cases where \gfour comes up with good (upper box) and bad (lower box) answers. This error is of the multi-hop dependency type. To make shortcakes, removing the cake from the oven (Step 10) is dependent on baking the cake (step 9) which is later dependent on combining the ingredients (Step 2). Examples of other error types can be found in \autoref{fig:lemon-cat-examples}.}
\label{fig:cherries-lemons}
\end{figure}

%When \llms are good, their generations can be really good. \nb{What is the point of the first sentence?}
%But it is more important to characterize their mistakes before they can be reliably used in critical settings.\nb{This sentence is also quite obvious. If you are looking to cut content for space, consider removing the above two sentences.}

To better understand model failures, we sampled and analyzed 50 explanations generated by \gfomni (A + E) where it produces an incorrect answer.
We identify 4 major types of errors:
\squishlist
\item Multi-hop dependency (40\%): Failure to understand that two steps might be related through an intermediate step. For instance, to make shortcakes in \autoref{fig:cherries-lemons}, removing the cake from the oven (Step 10) is dependent on baking the cake (Step 9) which, consequently, is dependent on combining the ingredients (Step 2).
The model does not seem to understand the transitive nature of such dependencies.
\item Effects (20\%): Failure to understand that an effect of the preceding step leads to the succeeding step, e.g., serving a cake in \autoref{fig:lemon-cat-examples} must happen after mixing ingredients and consequently baking.
\llms fail to identify additive effects of steps in a plan which enable a later step, leading to goal completion.
\item Preconditions (18\%): Failure to understand a condition that needs to be satisfied for a step to happen. For instance in \autoref{fig:lemon-cat-examples}, to add sauce in Step 20, meatballs need to cooked in Step 15 so they can be added to the sauce in Step 17.
\item Irrelevant Answers - Model produces answers that are unrelated to the step being asked about, e.g., in \autoref{fig:lemon-cat-examples} to make chocolate cake, the model's answer does not address a relevant step (Step 7) at all.
It is surprising to see that \llms mistakenly produce an answer about an unrelated step, particularly given that the input context is short (well below maximum context length) and can be easily used for grounding.
% \mooney{I'm not sure I would call this "hallucination" Why are you using this term here?}
\squishend

\section{Conclusion}

Understanding plans requires reasoning about its different aspects such as preconditions and effects.
% \nb{This is what Greg would call "content-free sentence" :-). Add "such as x, y, and z" or "which is difficult to do because..." etc.}
This paper introduces \ours, a new benchmark to evaluate the causal and temporal reasoning abilities about plans.
Despite the 
% \nb{need an adjective here}
remarkable strength of current SOTA \llms, we find that none of them are very good at understanding whether one step in a plan must precede (or succeed) another.
Particularly, they are much worse at knowing when there is {\it not} a dependency between steps.
We also find that \llm predictions are not robust as measured by two metrics of consistency.
Prompting \llms to provide an answer and then to explain it improves performance significantly, and is even better than chain-of-thought (reasoning followed by answering).
Human evaluation of these explanations shows that models have a long way to go at understanding dependencies.

\section*{Limitations}

While our work only considers cooking recipes as procedural texts, our methods can in principle be applied to many other domains.
Medical practice guidelines, repair manuals, and software tutorials among others are domains worth investigating.
Our work only investigates English-language documents and this limits the generalizability of our findings to other languages.
% \vc{This is what I would focus on, I think we can defend the cooking domain (well-represented in pretraining, has interesting non-dependent steps) while also acknowledging that it's limiting to only eval on one domain.}

We benchmark a reasonably diverse set of \llms.
Currently, we cover 3 model families and models of varying sizes.
Due to the current fast-paced landscape of \llm development, we will continue to evaluate more \llms on \ours.

% \vc{Other possible limitations include: didn't evaluate a ton of models, didn't evaluate human generated explanations, didn't include fine-tuning analysis.}

It is difficult for any one person to adequately evaluate the various aspects of plans, particularly recipes. 
To alleviate this problem, we use 3 crowd-sourced annotators to judge model explanations and consider their average judgment \cite{lal-etal-2022-using}, but recognize the limitations of this solution.
We do show high inter-annotator agreement \cite{lal-etal-2021-tellmewhy} using Weighted Fleiss Kappa \cite{Marasini2016AssessingTI}, demonstrating the reliability of our results.
While human evaluation is expensive and time-consuming and the number of experiments per model balloons costs exponentially, it is critical for open-ended generation tasks.
We evaluate enough explanations to obtain statistically significant results.
% It would also be good to obtain judgments for how well humans explain dependencies in plans.

% \vc{I probably wouldn't point to the human eval method as a limitation, as we also claim high IAA. I think there's probably other things we can discuss.}
% \yash{i have gotten human eval is a weakness and mturk noise as weaknesses in reviews}

We use BM25 as a reasonable choice to find similar exemplars.
We acknowledge that there are more modern techniques for selecting in-context examples, but this step is not the focus of our current work.
We leave further exploration of exemplar selection methods to future work.
For a domain like recipe text where texts are long and less amenable to a single embedding vector approach, keyword-based retrieval such as BM25 is very effective.

Since there are no gold explanations, we cannot combine few-shot prompting and chain-of-thought (or answer then explain) settings for \gfomni.
Note that we also do not advocate for using gold explanations along with automatic metrics to judge model explanations due to established inadequacies in using automatic metrics for free-form generations.

Due to very strict rate limits on the recently released Gemini models, we are unable to analyze \gemadv through chain-of-thought and other prompting techniques.
For consistency, we analyze \gfomni since it has similar performance (A + E) on \ours.
% \vc{I probably also wouldn't consider this a limitation by itself. I would rather say the limitation is that we didn't try other retrieval strategies or methods. But we should defend the choice of BM25, especially for a domain like recipes where the texts are long (less amenable to single embedding vector search) and keyword-based retrieval is a reasonable approach.}

\section*{Ethical Considerations}

% However, in real-world environments, this assumption is unlikely to hold.
% Users can generate malicious procedures by providing such hints to \llms.

Prior work has shown that \llms exhibit various types of bias.
While they do not generate free-form language for our binary prediction task, it is possible, though highly unlikely, that biases explicitly come up in the explanations.
Deploying such unreliable models into critical infrastructure and relying on them for decisions can cause harm to users.

\section*{Acknowledgements}

This material is based on research that is supported in part by the Air Force Research Laboratory (AFRL), DARPA, for the KAIROS program under agreement number FA8750-19-2-1003 and in part by the National Science Foundation under the award IIS \#2007290.
This material is also based upon work supported by the DARPA's Perceptually-enabled Task Guidance (PTG) program under Contract No. HR001122C007.

\bibliography{anthology,custom}
% \bibliography{custom}

\appendix

\clearpage

\section{Benchmark Models}
\label{appsec:benchmark_models}

We provide details of each model we evaluate on \ours.
For in-context example selection for the \llms we utilize the rank\_BM25 library\footnote{\url{https://github.com/dorianbrown/rank_bm25}} available under the Apache 2.0 license.

\paragraph{\texttt{gpt-4o-2024-05-13}} accepts as input any combination of text, audio, image, and video and generates any combination of text, audio, and image outputs.
It is especially better at vision and audio understanding compared to existing models.

\paragraph{\texttt{gpt-3.5-turbo}} is an instruction-tuned pretrained language model that powered the original ChatGPT application.

\paragraph{\texttt{gpt-4-turbo-2024-04-09}} is the first turbo model in the GPT-4 series. It is more capable and cheaper than GPT-3.5, and supports a 128K context window.
It is possibly a 8x220B Mixture-of-Experts model.

\paragraph{\texttt{gemini-1.5-flash-latest}} is a version of \gemadv optimized for low latency and inference cost.

\paragraph{\texttt{gemini-1.5-pro-latest}} is built on a Mixture-of-Experts (MoE) architecture.
\gemadv a mid-size multimodal model, optimized for scaling across a wide-range of tasks, and also introduces a breakthrough experimental feature in long-context understanding.
It is difficult to perform a wide range of experiments with this model due to the imposed rate limits.

\paragraph{\texttt{gpt-4o-mini-2024-07-18}} has a context window of 128K tokens, supports up to 16K output tokens per request.
It surpasses \gfturbo and other small models on academic benchmarks across both textual intelligence and multimodal reasoning, and supports the same range of languages as \gfomni.

\paragraph{\texttt{gemini-1.0-pro-latest}} is built on top of Transformer decoders that are enhanced with improvements in architecture and model optimization to enable stable training at scale and optimized inference. 
They are trained to support 32k context length, employing efficient attention mechanisms (for e.g. multi-query attention).
Gemini models are trained to accommodate textual input interleaved with a wide variety of audio and visual inputs, and \gem is the mid-sized model in the series.

\paragraph{\texttt{claude-3-5-sonnet-20240620}} sets new industry benchmarks for graduate-level reasoning (GPQA), undergraduate-level knowledge (MMLU), and coding proficiency (HumanEval). 
It shows marked improvement in grasping nuance, humor, and complex instructions, and is exceptional at writing high-quality content with a natural, relatable tone. 

\paragraph{\texttt{Meta-Llama3-8B-Instruct}} is a standard decoder-only transformer architecture similar to its predecessor Llama2.
Compared to Llama2, \llama uses a tokenizer with a vocabulary of 128K tokens that encodes language much more efficiently, which leads to substantially improved model performance.
It also uses grouped query attention (GQA) and was trained on sequences of 8,192 tokens, using a mask to ensure self-attention does not cross document boundaries.
\llama is pretrained on over 15T tokens that were all collected from publicly available sources.

\section{Additional Analysis Results}
\label{appsec:analysis}

We use scikit-learn classification report to calculate precision, recall and F1 score per-class and as macro average.

\begin{figure}[!t]
\small
\centerline{
\fbox{
    \parbox{0.95\columnwidth}{
	{\bf Goal:} fusilli and meatballs \\ 
    {\bf Steps:}
    ...\\
    15. Cook the meatballs until browned on all sides.\\
    16. Drain on kitchen paper.\\
    17. Add the meatballs to the tomato sauce.\\
    ...\\
    20. Toss with 3 ladlefuls of sauce in serving bowl..\\
    ...
    \squishlist
    \item[{\bf Question:}] Explain why or why not Step 20 must happen after Step 15.
    \item[{\bf Model Answer:}] Step 20 can happen after Step 19 as it involves tossing the cooked pasta with the sauce, which is independent of browning the meatballs in Step 15.
    % \item[{\bf Human Score:}] 5.0
    \squishend
    
    \hrule
    \vspace{0.2cm}
    {\bf Goal:} chocolate cake \\ 
    {\bf Steps:}\\
    1. Preheat oven to 180C / Gas Mark 4.\\
    2. Put chocolate chips, coffee, water in bowl.\\
    3. Microwave until melted.\\
    ...\\
    7. Gradually beat in sugar till fluffy.\\
    8. Gradually add and beat in egg yolks.\\
    ...
    \squishlist
    \item[{\bf Question:}] Explain why or why not Step 3 must happen before Step 7.
    \item[{\bf Model Answer:}] The chocolate mixture needs to be cooled before it is added to the batter in Step 10.
    % \item[{\bf Human Score:}] 1.0
    \squishend

    \hrule
    \vspace{0.2cm}
    {\bf Goal:} almond and apple cake \\ 
    {\bf Steps:}...\\
    7. Add half the flour and half the milk.\\
    8. Use a wooden spoon to stir until well combined.\\
    ...\\
    13. Set aside to cool slightly before plating.\\
    14. Serve warm.\\
    ...
    \squishlist
    \item[{\bf Question:}] Explain why or why not Step 14 must happen after Step 8.
    \item[{\bf Model Answer:}] Step 14 is the final step and cannot happen after mixing the ingredients.
    % \item[{\bf Human Score:}] 1.0
    \squishend
    }
}}
\caption{Examples of types of \gfour errors. The top box contains an error related to misunderstanding preconditions, the second one about producing irrelevant answers and the last one about misunderstanding effects.}
\label{fig:lemon-cat-examples}
\end{figure}

% \autoref{tab:close} and \autoref{tab:far} presents the performance of all models on questions which ask about steps that are close to each other and far from each other respectively.

% \input{tables/far}
% \input{tables/close}

% \autoref{tab:before} and \autoref{tab:after} shows the performance of the models on question that ask about different temporal relations between pairs of steps, before and after respectively.

% \input{tables/before}
% \input{tables/after}

\subsection{Understanding Directional Dependencies}

\input{figs/before_after_diff}
% \nb{Is this really about temporal relation? In one sense, the order question is also about temporality of the steps. While we use before/after the ask is really about causal effects and preconditions/causes. Given this I am not sure if we should call this temporal. Instead, we could consider calling this section Reasoning about Causal Effects.}
Next, we study how models handle questions about different aspects of the same pair of steps.
Typically, questions about why a step must happen \textit{before} another require reasoning about preconditions and causes, while answering why a step must happen \textit{after} another requires understanding the effects of any performed actions.
\autoref{fig:before-after-diff} shows the difference in F1 score between answering and providing an explanation (A + E) and the answer-only setting (A) with different models for these questions. 
Adding explanations helps all models understand effects better (\textit{after}).
We hypothesize that this is because effects in recipes can be more immediate and hence, would be easier to understand.
The biggest gain is seen for \gemadv and \gfomni, while \llama and \gem do not improve a lot.
We note that explanations significantly hurt \gthree and \flash in understanding preconditions, which is unusual.
Similar to \textit{after} questions, \llama and \gem do not improve a lot with explanations on \textit{before} questions.

\subsection{Reasoning as a function of Step Distance}

\input{figs/far_close_diff}
Next, we study how the distance between the steps in question impacts model performance.
A question is said to be about \emph{close} steps $(step_i, step_j)$ if $(j-i)<3$, and \emph{distant} otherwise.
\autoref{fig:far-close-diff} shows the difference in F1 score between answering then explaining (A + E) and just answering (A) with different models as a function of step distance.
Generating explanations helps models reason about distant steps, with \gfomni and \gemadv receiving the greatest benefit.
However, they don't help understand dependencies between close steps (which are easier to reason about).
In fact, producing explanations even hurts some models, particularly \gthree.
We hypothesize that models are likely to predict a dependencies between steps that are distant from each other, since it is likely that steps towards the end of a plan depend on ones near the start.
We find that this is indeed true, the recall for the nondependent is very low (usually, $\sim$20\%) and they are biased towards predicting a dependency between distant steps.

\subsection{Prompting Variations}

\autoref{tab:prompt_variants_full} presents more results for prompting techniques.
In particular, we show more variations of self-consistency.
We find that varying the temperature and number of samples does not make a significant difference.

\input{tables/gpt4o_variants}

\subsection{Error Examples}

We present examples of different types of errors in \autoref{fig:cherries-lemons} and \autoref{fig:lemon-cat-examples}.

\section{Human Evaluation Details}
\label{appsec:human_eval}

\subsection{Task Details}

\begin{figure}[!thb]
    \centering
    \includegraphics[width=\columnwidth]{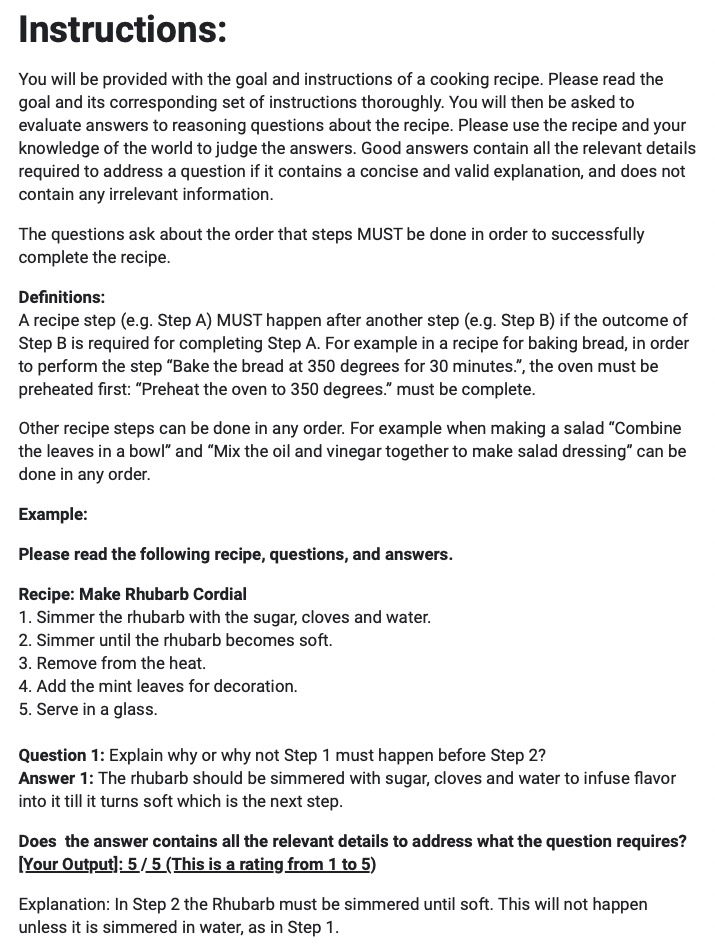}
    \caption{Instructions provided to annotators when making judgments about explanations for \dep questions.}
    \label{fig:d_instr}
\end{figure}

\begin{figure}[!thb]
    \centering
    \includegraphics[width=\columnwidth]{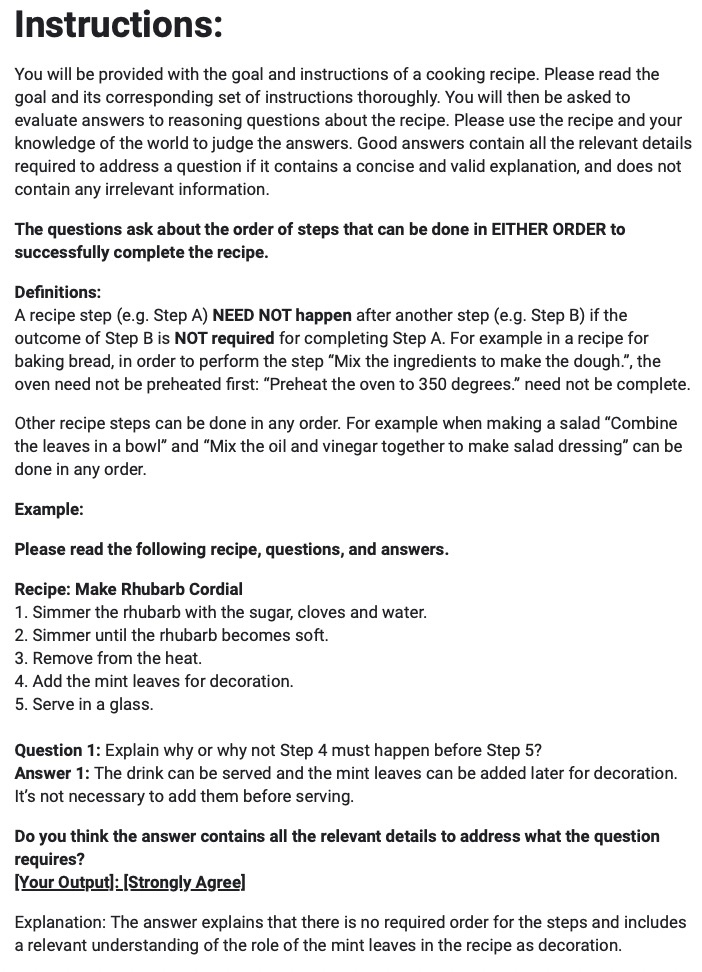}
    \caption{Instructions provided to annotators when making judgments about explanations for \ndep questions.}
    \label{fig:nd_instr}
\end{figure}

\autoref{fig:d_instr} and \autoref{fig:nd_instr} show the instructions as well as one of the examples presented to annotators when eliciting judgments for model explanations for \dep and \ndep questions.
For each HIT, workers are asked to read the goal of the plan and its steps and then evaluate 6 randomized questions and corresponding answers from models, providing judgments on a Likert scale of 1 to 5.
We only select US-based master turkers who have a minimum lifetime approval rating of 95\%.
On average, workers took 3 minutes and 51 seconds to judge 6 answers to questions about a plan.
We pay them \$1.5 per HIT which translates to \$23.35 per hour, significantly higher than federal and local minimum wage.

\subsection{Additional Human Evaluation Results}

We also used two additional metrics to interpret human judgments of model answers.
For \avgbin, we transform each score into a binary value (1 if \textgreater 3 and 0 otherwise), calculate the mean of these values for an answer and average them over all the data points.
We calculate the majority binary class of judgments for each explanation as \maj.
We report all the metrics for \dep and \ndep in \autoref{tab:d_human} and \autoref{tab:nd_human} respectively.
Looking at \avgbin, we note that there is room for improvement on the quality of model explanations.
\maj indicates that model explanations are convincing even when they are wrong.
Note that these metrics do not account for corresponding answer correctness for a model's explanation.
% \mooney{What additional conclusions can we draw from these additional metrics?}

\autoref{fig:histogram} presents the distribution of human judgment scores for explanations generated by various models (A + E). 
We note that models frequently produce high quality answers (5); however, they make too many errors (\textless 3 out of 5) to be consistently reliable .

\input{tables/d_humaneval}
\input{tables/nd_humaneval}

\input{figs/histogram}

\subsection{Inter-annotator Agreement}
\label{subsec:fleiss}

We measured the inter-rater reliability of annotators' judgments using weighted Fleiss’s Kappa \cite{Marasini2016AssessingTI}, following the weighting scheme used by \citet{bastan-etal-2020-authors}.
This measure has a penalty for each dissimilar rating based on the distance between the two ratings.
For instance, if two annotators classify a document as a positive, the agreement weight is 1, but if one classifies as a positive, and the other classifies as slightly positive the agreement weight is less.
The weights between different classes are shown in \autoref{tab:fleiss_kappa_weights} where negative, slightly negative, neutral, slightly positive, and positive classes are shown with -2, -1, 0, 1, and 2.
We follow the setup used in \citet{bastan-etal-2020-authors} for a similar multi-class labeling task.
% \mooney{You don't explain "binarized" version}

\input{tables/fleiss_weights}

\autoref{tab:fleiss} presents the inter-annotator agreement for judgments on answers to different types of questions in \ours.
The high Fleiss Kappa values demonstrate strong agreement between annotators and indicate reliability of our human evaluation framework.

\input{tables/fleiss_results}

\section{Prompts Used}
\label{appsec:prompts}

We present the different prompts used with the benchmark models in \autoref{fig:prompts}.
All models use the answer-only (A) prompt.
All models also share the (A + E) prompt except the Gemini models which use the NL (A + E) prompt instead.
We found that Gemini was better at producing free-form natural language as opposed to a structured code format.

\input{figs/prompt_fig}

We used a temperature of 0.0 for all the experiments with each model to select the most likely token at each step, as this setting allow for reproducibility\footnote{We note that some researchers have shown that even this setting might not make it completely reproducible: \url{https://twitter.com/ofirpress/status/1542610741668093952?s=46&t=f9v5k9RzVKnTK1e0UyauOA}}. 

We use the following code snippet to query any OpenAI models.

{\small
\begin{verbatim}
import openai

client = OpenAI(api_key=config["OPENAI_API_KEY"])

response = client.chat.completions.create(
                model=openai_model_name,
                messages=prompt,
                temperature=0.0,
                max_tokens=2,
                top_p=1.0,
                frequency_penalty=0.0,
                presence_penalty=0.0
            )
\end{verbatim}
}

We use the following code snippet to query any Gemini models.

{\small
\begin{verbatim}
import google.generativeai as genai

genai.configure(api_key=config["GEMINI_API_KEY"])

model = genai.GenerativeModel(args.model_name)
candidatecount, temp, topp, topk = 1, 0.0, 1.0, 1
generation_config = genai.GenerationConfig(
    candidate_count = candidatecount,
    max_output_tokens = args.max_tokens,
    temperature = temp,
    top_p = topp,
    top_k = topk
)
response = model.generate_content(prompt)
\end{verbatim}
}

We run inference on \llama locally on one 40GB Nvidia A6000 GPU using HuggingFace \cite{wolf-etal-2020-transformers}.

\end{document}

%% file: tables/prf.tex
\begin{table*}
\centering
\begin{tabular}{| l | c | c | c | c | c | c | c | c | c | c |}
\hline
 &  & \multicolumn{3}{c|}{\dep} & \multicolumn{3}{c|}{\ndep} & \multicolumn{3}{c|}{Macro Avg} \\
\hline
 &  & P & R & F & P & R & F & P & R & F \\
\hline
\multirow{2}{*}{\gthree}  & (A) & 0.62 & 0.50& 0.55 & 0.58 & \textbf{0.69} & 0.63 & 0.60& 0.60& 0.59 \\
 & (A+E) & 0.56 & 0.71 & 0.63 & 0.61 & 0.45 & 0.52 & 0.58 & 0.58 & 0.57 \\
\hline
\multirow{2}{*}{\gfturbo} & (A) & 0.57 & 0.81 & 0.67 & 0.67 & 0.39 & 0.49 & 0.62 & 0.60& 0.58 \\
 & (A+E) & 0.66 & 0.79 & 0.72 & 0.74 & 0.59 & 0.66 & 0.70& 0.69 & 0.69 \\
\hline
\multirow{2}{*}{\gfomni} & (A) & 0.53 & 0.92 & 0.67 & 0.71 & 0.19 & 0.30 & 0.62 & 0.55 & 0.49 \\
 & (A+E) & 0.66 & 0.86 & 0.75 & 0.80 & 0.57 & 0.66 & 0.73 & 0.71 & 0.70\\
\hline
\multirow{2}{*}{\gfmini} & (A) & 0.53 & 0.88 & 0.76 & 0.64 & 0.22 & 0.33 & 0.59 & 0.55 & 0.50 \\
 & (A+E) & 0.62 & 0.78 & 0.69 & 0.70 & 0.52 & 0.59 & 0.66 & 0.65 & 0.64 \\
\hline
\multirow{2}{*}{\llama} & (A) & 0.52 & 0.84 & 0.64 & 0.59 & 0.23 & 0.33 & 0.56 & 0.54 & 0.49 \\
 & (A+E) & 0.53 & 0.82 & 0.64 & 0.59 & 0.26 & 0.36 & 0.56 & 0.54 & 0.50\\
\hline
\multirow{2}{*}{\gem} & (A) & 0.57 & 0.45 & 0.50& 0.55 & 0.66 & 0.60& 0.56 & 0.55 & 0.55 \\
 & (A+E) & 0.56 & 0.65 & 0.60& 0.59 & 0.50& 0.54 & 0.58 & 0.57 & 0.57 \\
\hline
\multirow{2}{*}{\gemadv} & (A) & 0.55 & 0.77 & 0.64 & 0.61 & 0.37 & 0.46 & 0.58 & 0.57 & 0.55 \\
 & (A+E) & \textbf{0.67} & 0.93 & \textbf{0.78} & 0.89 & 0.54 & \textbf{0.67} & \textbf{0.78} & \textbf{0.74} & \textbf{0.73} \\
\hline
\multirow{2}{*}{\flash} & (A) & 0.55 & 0.69 & 0.61 & 0.58 & 0.43 & 0.49 & 0.56 & 0.56 & 0.55 \\
 & (A+E) & 0.54 & 0.75 & 0.63 & 0.59 & 0.37 & 0.46 & 0.57 & 0.56 & 0.54 \\
\hline
\multirow{2}{*}{\sonnet} & (A) & 0.58 & 0.76 & 0.66 &  0.65 & 0.46 & 0.54 & 0.62 & 0.61 & 0.60 \\
 & (A+E) & 0.63 & \textbf{0.97} & 0.76 & \textbf{0.93} & 0.44 & 0.60 & \textbf{0.78} & 0.70 & 0.68 \\
\hline
\end{tabular}
\caption{
Performance of all models on \binarytask~ when just providing an answer (A) and when also explaining that answer (A+E).
We report per-label as well as macro average precision, recall and F1 score.
% \vc{remove quadrant of lines in upper left corner}
% \yash{tried but there is always a line that comes in between}
% \gemadv is the best model.
}
\label{tab:auto_eval}
\end{table*}

%% file: tables/humaneval.tex
\begin{table}[!ht]
\centering
\begin{tabular}{| l | r | r |}
\hline
 & \avg & \modavg \\
\hline
\llama & 3.26 & 1.87 \\
\hline
\gfturbo & 3.85 & 2.90 \\
\hline
\gfomni & 3.84 & 2.93 \\
\hline
\gemadv & 3.83 & 2.69 \\
\hline
\end{tabular}
\caption{Human evaluation metrics for explanations generated by various models in the (A+E) setting.}
\label{tab:all_human}
\end{table}

%% file: tables/consistency.tex
\begin{table}
\centering
\begin{tabular}{| l | >{\small}c | >{\small}c |}
\hline
 & \tc & \ndsc \\
\hline
\gthree (A) & 52.39\% & 70.42\% \\
\hline
\gthree (A+E) & 49.23\% & 73.31\% \\
\hline
\gfturbo (A) & 48.87\% & 70.28\% \\
\hline
\gfturbo (A+E) & 55.00\% & 66.97\% \\
\hline
% \gfturbo (E$\rightarrow$P) & 74.93\% & 67.75\% \\
% \hline
\gfomni (A) & \textbf{79.86\%} & 47.96\% \\
\hline
\gfomni (A+E) & 67.46\% & 58.17\% \\
\hline
% \gfomni (E$\rightarrow$P) & 83.66\% & 53.73\% \\
% \hline
\gfmini (A) & 70.56\% & 54.79\% \\
\hline
\gfmini (A+E) & 57.54\% & 56.97\% \\
\hline
\llama (A) & 60.42\% & \textbf{83.87\%} \\
\hline
\llama (A+E) & 55.77\% & 83.38\% \\
\hline
\gem (A) & 53.38\% & 73.80\% \\
\hline
\gem (A+E) & 49.79\% & 66.90\% \\
\hline
\gemadv (A) & 55.14\% & 58.24\% \\
\hline
\gemadv (A+E) & 79.65\% & 60.21\% \\
\hline
\flash (A) & 45.92\% & 79.44\% \\
\hline
\flash (A+E) & 43.10\% & 76.62\% \\
\hline
\sonnet (A) & 45.14\% & 50.21\% \\
\hline
\sonnet (A+E) & 76.83\% & 48.10\% \\
\hline
\end{tabular}
\caption{Robustness of different models on two consistency metrics, \tc and \ndsc.}
\label{tab:robustness}
\end{table}

%% file: tables/cot.tex
\begin{table}[!htb]
\centering
\begin{tabular}{| c | c | c | c | c |}
\hline
 & P & R & F1 & \tc \\
\hline
(A) & 0.62 & 0.55 & 0.49 & 79.86\% \\
\hline
(E+A) & 0.77 & 0.65 & 0.6 & 83.66\% \\
\hline
(A+E) & 0.73 & 0.71 & 0.7 & 67.46\% \\
\hline
\end{tabular}
\caption{Performance of \gfomni on the Step Order Prediction task when just predicting the dependency (A) vs predicting and explaining the judgment (A+E) vs using chain-of-thought prompting (E+A).}
\label{tab:binary_explain_vs_cot}
\end{table}

%% file: tables/d_cot.tex
\begin{table}[!ht]
\centering
\begin{tabular}{| c | c | c | c |}
\hline
 & P & R & F1 \\
\hline
(A) & 0.53 & 0.92 & 0.67 \\
\hline
(E+A) & 0.59 & 0.98 & 0.74 \\
\hline
(A+E) & 0.66 & 0.86 & 0.75 \\
\hline
\end{tabular}
\caption{Performance of \gfomni on the Step Order Prediction task when just predicting the dependency (A) vs predicting and explaining the judgment (A+E) vs using chain-of-thought prompting (E+A) on \dep.}
\label{tab:d_cot}
\end{table}

%% file: tables/prompting.tex
\begin{table}[!ht]
\centering
\begin{tabular}{| l | l | l | l |}
\hline
 & P & R & F \\
\hline
Zero-shot  & 0.62 & 0.55 & 0.49 \\
\hline
Self-Consistency & 0.62 & 0.55 & 0.49 \\
\hline
Few-shot & 0.79 & 0.70 & 0.68 \\
\hline
\end{tabular}
\caption{Performance of different prompting techniques with \gfomni on Step Order Prediction. For self-consistency, we report $k=3$ and $temp=0.6$ here, and use 5 exemplars for few-shot experiments.}
\label{tab:prompt_variants}
\end{table}

%% file: figs/before_after_diff.tex
\begin{figure}[!ht]
    \centering
    \includegraphics[width=\columnwidth]{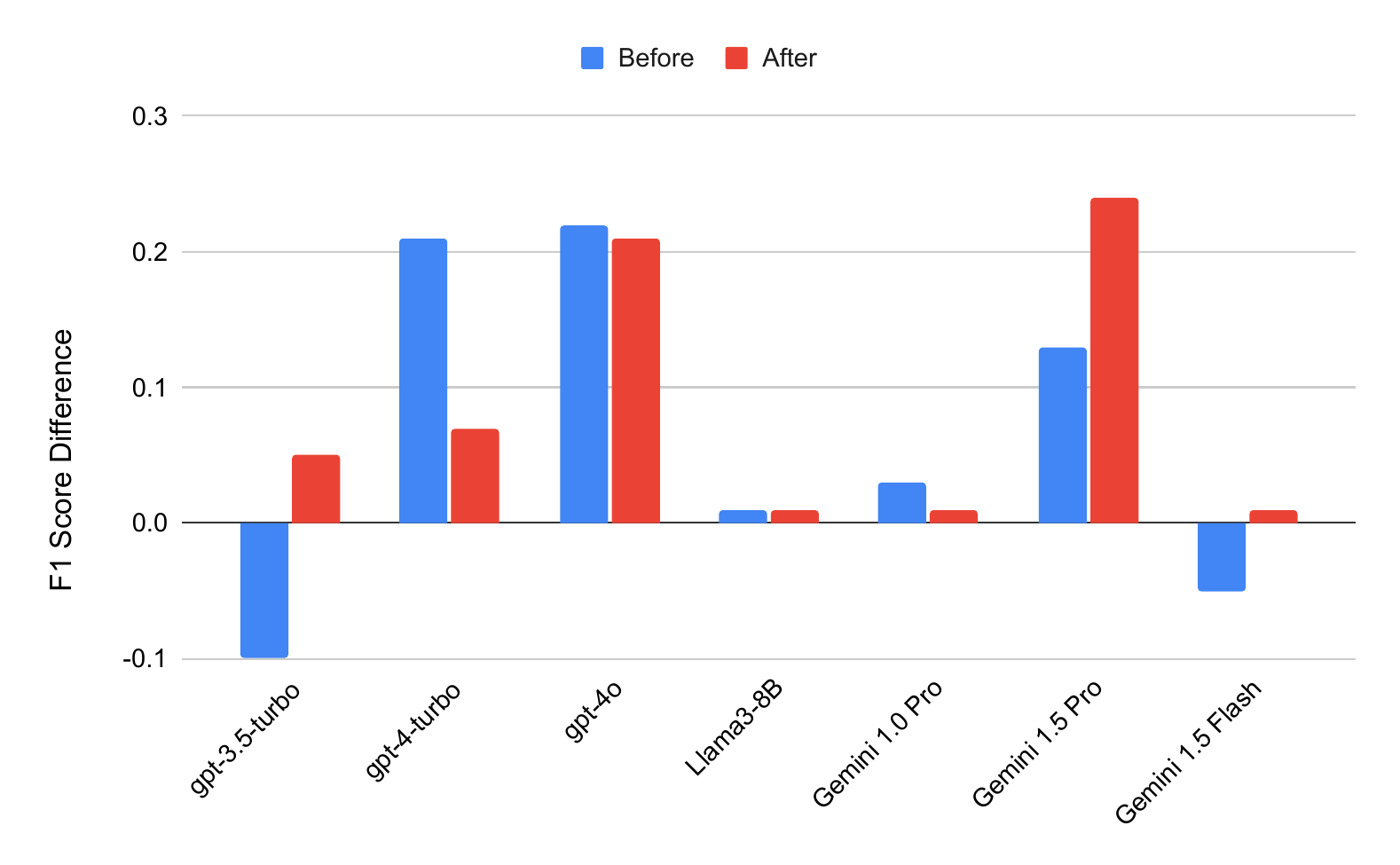}
    \caption{Difference in  model performance between (A+E) and (A) settings split by temporal relation type (\textit{before} and \textit{after}) asked about in the question. We subtract F1 score in the (A) from the (A+E) setting.}
    \label{fig:before-after-diff}
\end{figure}

%% file: figs/far_close_diff.tex
\begin{figure}[!ht]
    \centering
    \includegraphics[width=\columnwidth]{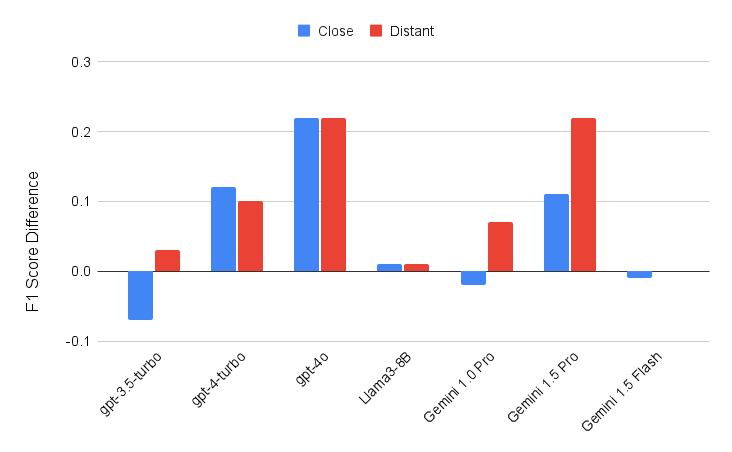}
    \caption{Difference in performance of models between (A+E) and (A) settings split by the distance between the steps being asked about in the question.}
    \label{fig:far-close-diff}
\end{figure}

%% file: tables/gpt4o_variants.tex
\begin{table*}[!tbh]
\centering
\begin{tabular}{| l | c | c | c | c | c |}
\hline
 & P & R & F & \tc & \ndsc \\
\hline
Zero-shot  & 0.62 & 0.55 & 0.49 & 79.86\% & 47.96\% \\
\hline
Self-Consistency [3, 0.6]  & 0.62 & 0.55 & 0.49 & 78.66\% & 48.17\% \\
Self-Consistency [3, 0.8]  & 0.61 & 0.55 & 0.48 & 79.58\% & 47.25\% \\
Self-Consistency [5, 0.6]  & 0.61 & 0.55 & 0.48 & 79.72\% & 46.76\% \\
Self-Consistency [5, 0.8]  & 0.61 & 0.55 & 0.48 & 79.86\% & 45.99\% \\
\hline
Few-shot & 0.79 & 0.70 & 0.68 & 81.48\% & 45.56\% \\
\hline
\end{tabular}
\caption{Accuracy and consistency of different prompting techniques with \gfomni on the Step Order Prediction Task. The first number within the square bracket in self-consistency experiments represents the number of samples and the second represents the temperature at which predictions were sampled. For few-shot experiments, we use k=5 exemplars and use BM25 \cite{bm25} to dynamically retrieve exemplars from a held-out set that are closest to the test instance.}
\label{tab:prompt_variants_full}
\end{table*}

%% file: tables/d_humaneval.tex
\begin{table*}[!tbh]
\centering
\begin{tabular}{| l | r | r | r | r |}
\hline
 & \avg & \modavg & \avgbin & \maj \\
\hline
\llama & 3.77 & 3.14 & 0.70 & 75.42 \\
\hline
\gfturbo & 3.95 & 3.39 & 0.77 & 79.58 \\
\hline
\gfomni & 4.08 & 3.58 & 0.83 & 87.92 \\
\hline
\gemadv & 3.98 & 3.8 & 0.77 & 87.08 \\
\hline
\end{tabular}
\caption{Human evaluation metrics for explanations generated by various models for \dep questions.}
\label{tab:d_human}
\end{table*}

%% file: tables/nd_humaneval.tex
% \begin{table}[!ht]
\begin{table*}[!tbh]
\centering
\begin{tabular}{| l | r | r | r | r |}
\hline
 & \avg & \modavg & \avgbin & \maj \\
\hline
\llama & 2.75 & 0.6 & 0.40 & 35.83 \\
\hline
\gfturbo & 3.74 & 2.41 & 0.70 & 75.83 \\
\hline
\gfomni & 3.6 & 2.29 & 0.66 & 71.67 \\
\hline
\gemadv & 3.69 & 1.58 & 0.68 & 75.42 \\
\hline
\end{tabular}
\caption{Human evaluation metrics for explanations generated by various models for \ndep questions.}
\label{tab:nd_human}
\end{table*}

%% file: figs/histogram.tex
\begin{figure}[!ht]
    \centering
    \includegraphics[width=\columnwidth]{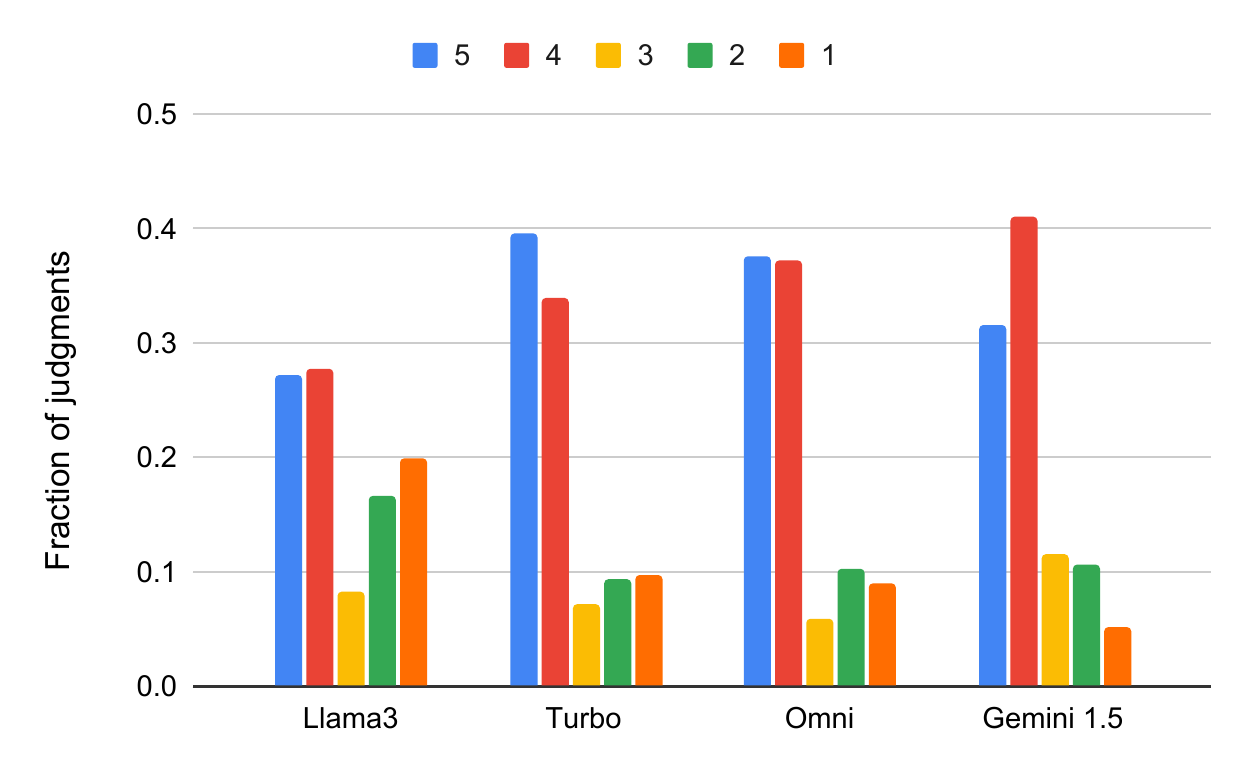}
    \caption{Distribution of human judgment scores for explanations generated by various models (A+E).}
    \label{fig:histogram}
\end{figure}

%% file: tables/fleiss_weights.tex
\begin{table}[!tbh]
    \small
    \centering
    \begin{tabular}{|c|c|c|c|c|c|}
    \toprule
         & -2 & -1 & 0 & 1 & 2 \\
        \midrule
        -2 & 1 & cos $\pi$/8 & cos $\pi$/4 & cos 3$\pi$/8 & 0 \\
        -1 & cos $\pi$/8 & 1 & cos $\pi$/8 & cos $\pi$/4 & cos 3$\pi$/8 \\
        0 & cos $\pi$/4 & cos $\pi$/8 & 1 & cos $\pi$/8 & cos $\pi$/4 \\
        1 & cos 3$\pi$/8 & cos $\pi$/4 & cos $\pi$/8 & 1 & cos $\pi$/8 \\
        2 & 0 & cos 3$\pi$/8 & cos $\pi$/4 & cos $\pi$/8 & 1 \\
    \bottomrule
    \end{tabular}
    \caption{Inter-class weights used for computing inter-annotator agreement}
    \label{tab:fleiss_kappa_weights}
\end{table}

%% file: tables/fleiss_results.tex
\begin{table}[!ht]
\small
\centering
\begin{tabular}{| l | r | r |}
\toprule
 & \makecell{Weighted Fleiss \\ Kappa} & \makecell{Weighted Binarized \\ Fleiss Kappa} \\
\midrule
\dep & 0.808 & 0.941 \\
\ndep & 0.705 & 0.934 \\
\bottomrule
\end{tabular}
\caption{Inter-annotator agreement as measured by Fleiss Kappa for each question type in \ours}
\label{tab:fleiss}
\end{table}

%% file: figs/prompt_fig.tex
\begin{figure*}[!tbh]
    \centering
    \includegraphics[width=\textwidth]{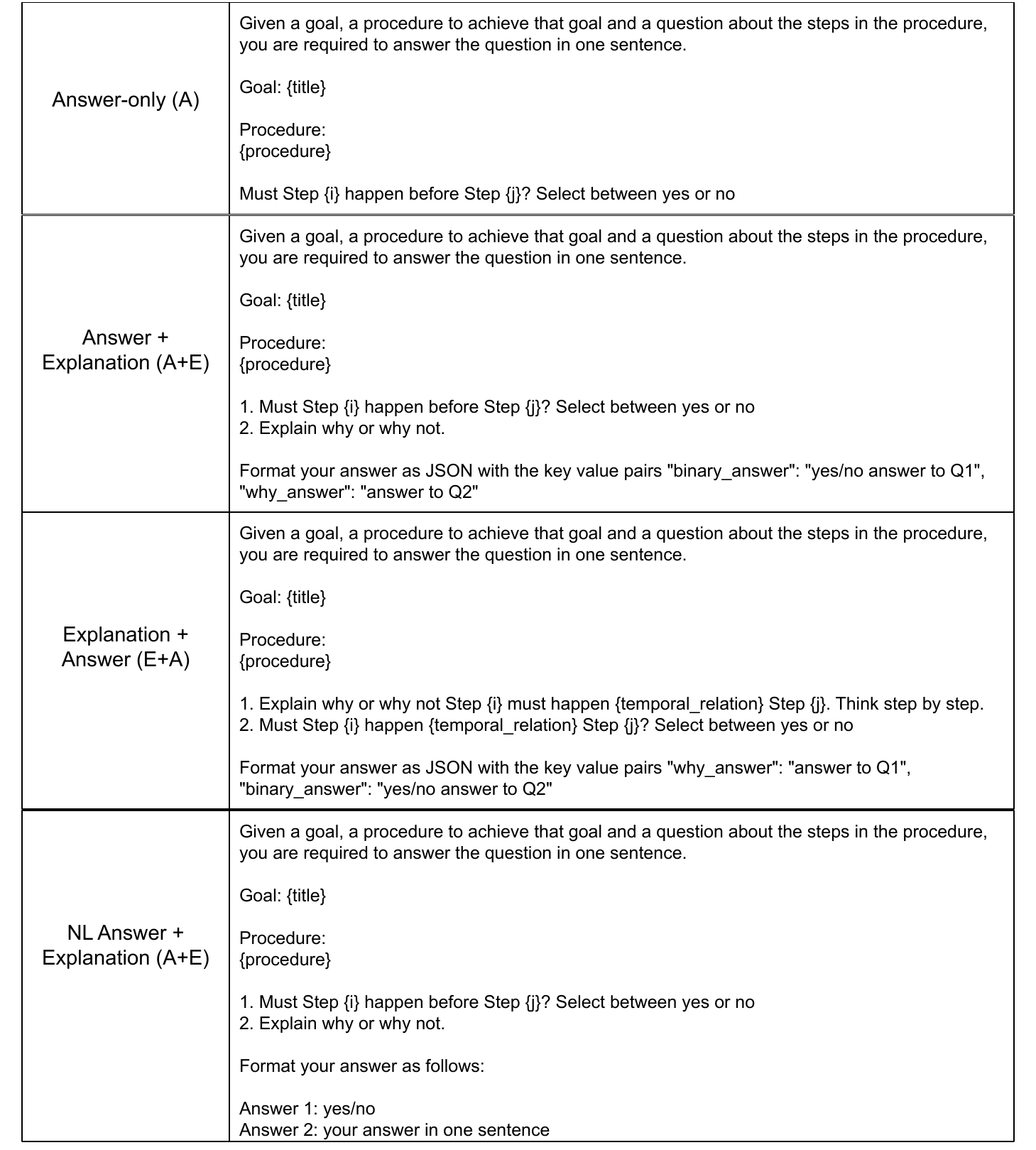}
    \caption{Different prompts used for our experiment settings and models. $i$ and $j$ represent step numbers and $temporal\_relation$ can be before/after.}
    \label{fig:prompts}
\end{figure*}

%% file: acl_latex.bbl
\begin{thebibliography}{54}
\providecommand{\natexlab}[1]{#1}

\bibitem[{Aouladomar and Saint-Dizier(2005)}]{aouladomar-saint-dizier-2005-towards}
Farida Aouladomar and Patrick Saint-Dizier. 2005.
\newblock \href {https://aclanthology.org/W05-1618} {Towards generating procedural texts: An exploration of their rhetorical and argumentative structure}.
\newblock In \emph{Proceedings of the Tenth {E}uropean Workshop on Natural Language Generation ({ENLG}-05)}, Aberdeen, Scotland. Association for Computational Linguistics.

\bibitem[{Bastan et~al.(2020)Bastan, Koupaee, Son, Sicoli, and Balasubramanian}]{bastan-etal-2020-authors}
Mohaddeseh Bastan, Mahnaz Koupaee, Youngseo Son, Richard Sicoli, and Niranjan Balasubramanian. 2020.
\newblock \href {https://doi.org/10.18653/v1/2020.coling-main.52} {Author{'}s sentiment prediction}.
\newblock In \emph{Proceedings of the 28th International Conference on Computational Linguistics}, pages 604--615, Barcelona, Spain (Online). International Committee on Computational Linguistics.

\bibitem[{Bosselut et~al.(2018)Bosselut, Levy, Holtzman, Ennis, Fox, and Choi}]{Bosselut2018SimulatingAD}
Antoine Bosselut, Omer Levy, Ari Holtzman, Corin Ennis, Dieter Fox, and Yejin Choi. 2018.
\newblock Simulating action dynamics with neural process networks.
\newblock \emph{ICLR}.

\bibitem[{Brahman et~al.(2023)Brahman, Bhagavatula, Pyatkin, Hwang, Li, Arai, Sanyal, Sakaguchi, Ren, and Choi}]{plasma}
Faeze Brahman, Chandra Bhagavatula, Valentina Pyatkin, Jena~D. Hwang, Xiang~Lorraine Li, Hirona~J. Arai, Soumya Sanyal, Keisuke Sakaguchi, Xiang Ren, and Yejin Choi. 2023.
\newblock \href {https://arxiv.org/abs/2305.19472} {Plasma: Making small language models better procedural knowledge models for (counterfactual) planning}.
\newblock \emph{Preprint}, arXiv:2305.19472.

\bibitem[{Brown et~al.(2020)Brown, Mann, Ryder, Subbiah, Kaplan, Dhariwal, Neelakantan, Shyam, Sastry, Askell, Agarwal, Herbert-Voss, Krueger, Henighan, Child, Ramesh, Ziegler, Wu, Winter, Hesse, Chen, Sigler, Litwin, Gray, Chess, Clark, Berner, McCandlish, Radford, Sutskever, and Amodei}]{brown-etal-2020-fewshot}
Tom Brown, Benjamin Mann, Nick Ryder, Melanie Subbiah, Jared~D Kaplan, Prafulla Dhariwal, Arvind Neelakantan, Pranav Shyam, Girish Sastry, Amanda Askell, Sandhini Agarwal, Ariel Herbert-Voss, Gretchen Krueger, Tom Henighan, Rewon Child, Aditya Ramesh, Daniel Ziegler, Jeffrey Wu, Clemens Winter, Chris Hesse, Mark Chen, Eric Sigler, Mateusz Litwin, Scott Gray, Benjamin Chess, Jack Clark, Christopher Berner, Sam McCandlish, Alec Radford, Ilya Sutskever, and Dario Amodei. 2020.
\newblock \href {https://proceedings.neurips.cc/paper_files/paper/2020/file/1457c0d6bfcb4967418bfb8ac142f64a-Paper.pdf} {Language models are few-shot learners}.
\newblock In \emph{Advances in Neural Information Processing Systems}, volume~33, pages 1877--1901. Curran Associates, Inc.

\bibitem[{Caglayan et~al.(2020)Caglayan, Madhyastha, and Specia}]{caglayan-etal-2020-curious}
Ozan Caglayan, Pranava Madhyastha, and Lucia Specia. 2020.
\newblock \href {https://doi.org/10.18653/v1/2020.coling-main.210} {Curious case of language generation evaluation metrics: A cautionary tale}.
\newblock In \emph{Proceedings of the 28th International Conference on Computational Linguistics}, pages 2322--2328, Barcelona, Spain (Online). International Committee on Computational Linguistics.

\bibitem[{Camburu et~al.(2018)Camburu, Rockt\"{a}schel, Lukasiewicz, and Blunsom}]{camburu-etal-2018-esnli}
Oana-Maria Camburu, Tim Rockt\"{a}schel, Thomas Lukasiewicz, and Phil Blunsom. 2018.
\newblock \href {https://proceedings.neurips.cc/paper_files/paper/2018/file/4c7a167bb329bd92580a99ce422d6fa6-Paper.pdf} {e-snli: Natural language inference with natural language explanations}.
\newblock In \emph{Advances in Neural Information Processing Systems}, volume~31. Curran Associates, Inc.

\bibitem[{Chen et~al.(2019)Chen, Stanovsky, Singh, and Gardner}]{chen-etal-2019-evaluating}
Anthony Chen, Gabriel Stanovsky, Sameer Singh, and Matt Gardner. 2019.
\newblock \href {https://doi.org/10.18653/v1/D19-5817} {Evaluating question answering evaluation}.
\newblock In \emph{Proceedings of the 2nd Workshop on Machine Reading for Question Answering}, pages 119--124, Hong Kong, China. Association for Computational Linguistics.

\bibitem[{Dalvi et~al.(2018)Dalvi, Huang, Tandon, Yih, and Clark}]{dalvi-etal-2018-tracking}
Bhavana Dalvi, Lifu Huang, Niket Tandon, Wen-tau Yih, and Peter Clark. 2018.
\newblock \href {https://doi.org/10.18653/v1/N18-1144} {Tracking state changes in procedural text: a challenge dataset and models for process paragraph comprehension}.
\newblock In \emph{Proceedings of the 2018 Conference of the North {A}merican Chapter of the Association for Computational Linguistics: Human Language Technologies, Volume 1 (Long Papers)}, pages 1595--1604, New Orleans, Louisiana. Association for Computational Linguistics.

\bibitem[{Dalvi et~al.(2019)Dalvi, Tandon, Bosselut, Yih, and Clark}]{dalvi-etal-2019-everything}
Bhavana Dalvi, Niket Tandon, Antoine Bosselut, Wen-tau Yih, and Peter Clark. 2019.
\newblock \href {https://doi.org/10.18653/v1/D19-1457} {Everything happens for a reason: Discovering the purpose of actions in procedural text}.
\newblock In \emph{Proceedings of the 2019 Conference on Empirical Methods in Natural Language Processing and the 9th International Joint Conference on Natural Language Processing (EMNLP-IJCNLP)}, pages 4496--4505, Hong Kong, China. Association for Computational Linguistics.

\bibitem[{Diallo et~al.(2024)Diallo, Bikakis, Dickens, Hunter, and Miller}]{diallo-etal-2024-pizzacommonsense}
Aissatou Diallo, Antonis Bikakis, Luke Dickens, Anthony Hunter, and Rob Miller. 2024.
\newblock \href {https://arxiv.org/abs/2401.06930} {Pizzacommonsense: Learning to model commonsense reasoning about intermediate steps in cooking recipes}.
\newblock \emph{Preprint}, arXiv:2401.06930.

\bibitem[{Donatelli et~al.(2021)Donatelli, Schmidt, Biswas, K{\"o}hn, Zhai, and Koller}]{donatelli-etal-2021-aligning}
Lucia Donatelli, Theresa Schmidt, Debanjali Biswas, Arne K{\"o}hn, Fangzhou Zhai, and Alexander Koller. 2021.
\newblock \href {https://doi.org/10.18653/v1/2021.emnlp-main.554} {Aligning actions across recipe graphs}.
\newblock In \emph{Proceedings of the 2021 Conference on Empirical Methods in Natural Language Processing}, pages 6930--6942, Online and Punta Cana, Dominican Republic. Association for Computational Linguistics.

\bibitem[{Elazar et~al.(2021)Elazar, Kassner, Ravfogel, Ravichander, Hovy, Sch{\"u}tze, and Goldberg}]{elazar-etal-2021-measuring}
Yanai Elazar, Nora Kassner, Shauli Ravfogel, Abhilasha Ravichander, Eduard Hovy, Hinrich Sch{\"u}tze, and Yoav Goldberg. 2021.
\newblock \href {https://doi.org/10.1162/tacl_a_00410} {Measuring and improving consistency in pretrained language models}.
\newblock \emph{Transactions of the Association for Computational Linguistics}, 9:1012--1031.

\bibitem[{Henaff et~al.(2017)Henaff, Weston, Szlam, Bordes, and LeCun}]{henaff2017tracking}
Mikael Henaff, Jason Weston, Arthur Szlam, Antoine Bordes, and Yann LeCun. 2017.
\newblock Tracking the world state with recurrent entity networks.
\newblock \emph{ICLR}.

\bibitem[{Hou et~al.(2023)Hou, Zhang, and Callison-Burch}]{choice75}
Zhaoyi~Joey Hou, Li~Zhang, and Chris Callison-Burch. 2023.
\newblock Choice-75: A dataset on decision branching in script learning.
\newblock \emph{arXiv preprint arXiv:2309.11737}.

\bibitem[{Howcroft et~al.(2020)Howcroft, Belz, Clinciu, Gkatzia, Hasan, Mahamood, Mille, van Miltenburg, Santhanam, and Rieser}]{howcroft-etal-2020-twenty}
David~M. Howcroft, Anya Belz, Miruna-Adriana Clinciu, Dimitra Gkatzia, Sadid~A. Hasan, Saad Mahamood, Simon Mille, Emiel van Miltenburg, Sashank Santhanam, and Verena Rieser. 2020.
\newblock \href {https://doi.org/10.18653/v1/2020.inlg-1.23} {Twenty years of confusion in human evaluation: {NLG} needs evaluation sheets and standardised definitions}.
\newblock In \emph{Proceedings of the 13th International Conference on Natural Language Generation}, pages 169--182, Dublin, Ireland. Association for Computational Linguistics.

\bibitem[{Jiang et~al.(2019)Jiang, Zhang, Khandelwal, and Stone}]{jiang2019task}
Yuqian Jiang, Shiqi Zhang, Piyush Khandelwal, and Peter Stone. 2019.
\newblock \href {https://arxiv.org/abs/1804.08229} {Task planning in robotics: an empirical comparison of pddl-based and asp-based systems}.
\newblock \emph{Preprint}, arXiv:1804.08229.

\bibitem[{Kiddon et~al.(2015)Kiddon, Ponnuraj, Zettlemoyer, and Choi}]{kiddon-etal-2015-mise}
Chlo{\'e} Kiddon, Ganesa~Thandavam Ponnuraj, Luke Zettlemoyer, and Yejin Choi. 2015.
\newblock \href {https://doi.org/10.18653/v1/D15-1114} {Mise en place: Unsupervised interpretation of instructional recipes}.
\newblock In \emph{Proceedings of the 2015 Conference on Empirical Methods in Natural Language Processing}, pages 982--992, Lisbon, Portugal. Association for Computational Linguistics.

\bibitem[{Kumar and Talukdar(2020)}]{kumar-talukdar-2020-nile}
Sawan Kumar and Partha Talukdar. 2020.
\newblock \href {https://doi.org/10.18653/v1/2020.acl-main.771} {{NILE} : Natural language inference with faithful natural language explanations}.
\newblock In \emph{Proceedings of the 58th Annual Meeting of the Association for Computational Linguistics}, pages 8730--8742, Online. Association for Computational Linguistics.

\bibitem[{Lal et~al.(2021)Lal, Chambers, Mooney, and Balasubramanian}]{lal-etal-2021-tellmewhy}
Yash~Kumar Lal, Nathanael Chambers, Raymond Mooney, and Niranjan Balasubramanian. 2021.
\newblock \href {https://doi.org/10.18653/v1/2021.findings-acl.53} {{T}ell{M}e{W}hy: A dataset for answering why-questions in narratives}.
\newblock In \emph{Findings of the Association for Computational Linguistics: ACL-IJCNLP 2021}, pages 596--610, Online. Association for Computational Linguistics.

\bibitem[{Lal et~al.(2022)Lal, Tandon, Aggarwal, Liu, Chambers, Mooney, and Balasubramanian}]{lal-etal-2022-using}
Yash~Kumar Lal, Niket Tandon, Tanvi Aggarwal, Horace Liu, Nathanael Chambers, Raymond Mooney, and Niranjan Balasubramanian. 2022.
\newblock \href {https://doi.org/10.18653/v1/2022.emnlp-main.79} {Using commonsense knowledge to answer why-questions}.
\newblock In \emph{Proceedings of the 2022 Conference on Empirical Methods in Natural Language Processing}, pages 1204--1219, Abu Dhabi, United Arab Emirates. Association for Computational Linguistics.

\bibitem[{Lal et~al.(2024)Lal, Zhang, Brahman, Majumder, Clark, and Tandon}]{lal-etal-2024-tailoring}
Yash~Kumar Lal, Li~Zhang, Faeze Brahman, Bodhisattwa~Prasad Majumder, Peter Clark, and Niket Tandon. 2024.
\newblock \href {https://doi.org/10.18653/v1/2024.findings-acl.921} {Tailoring with targeted precision: Edit-based agents for open-domain procedure customization}.
\newblock In \emph{Findings of the Association for Computational Linguistics ACL 2024}, pages 15597--15611, Bangkok, Thailand and virtual meeting. Association for Computational Linguistics.

\bibitem[{LaValle(2006)}]{planning-algo-book}
Steven~M. LaValle. 2006.
\newblock \emph{Planning Algorithms}.
\newblock Cambridge University Press, USA.

\bibitem[{Le et~al.(2023)Le, Guo, Xu, and Ritter}]{chattychef}
Duong~Minh Le, Ruohao Guo, Wei Xu, and Alan Ritter. 2023.
\newblock \href {https://arxiv.org/abs/2305.17280} {Improved instruction ordering in recipe-grounded conversation}.
\newblock \emph{Preprint}, arXiv:2305.17280.

\bibitem[{Likert(1932)}]{likert1932technique}
Rensis Likert. 1932.
\newblock A technique for the measurement of attitudes.
\newblock \emph{Archives of psychology}.

\bibitem[{Lin et~al.(2020)Lin, Rao, Celikyilmaz, Nouri, Brockett, Dey, and Dolan}]{lin-etal-2020-recipe}
Angela Lin, Sudha Rao, Asli Celikyilmaz, Elnaz Nouri, Chris Brockett, Debadeepta Dey, and Bill Dolan. 2020.
\newblock \href {https://doi.org/10.18653/v1/2020.acl-main.440} {A recipe for creating multimodal aligned datasets for sequential tasks}.
\newblock In \emph{Proceedings of the 58th Annual Meeting of the Association for Computational Linguistics}, pages 4871--4884, Online. Association for Computational Linguistics.

\bibitem[{Lyu et~al.(2021)Lyu, Zhang, and Callison-Burch}]{lyu-etal-2021-goal}
Qing Lyu, Li~Zhang, and Chris Callison-Burch. 2021.
\newblock \href {https://doi.org/10.18653/v1/2021.inlg-1.19} {Goal-oriented script construction}.
\newblock In \emph{Proceedings of the 14th International Conference on Natural Language Generation}, pages 184--200, Aberdeen, Scotland, UK. Association for Computational Linguistics.

\bibitem[{Ma et~al.(2019)Ma, Wei, Bojar, and Graham}]{ma-etal-2019-results}
Qingsong Ma, Johnny Wei, Ond{\v{r}}ej Bojar, and Yvette Graham. 2019.
\newblock \href {https://doi.org/10.18653/v1/W19-5302} {Results of the {WMT}19 metrics shared task: Segment-level and strong {MT} systems pose big challenges}.
\newblock In \emph{Proceedings of the Fourth Conference on Machine Translation (Volume 2: Shared Task Papers, Day 1)}, pages 62--90, Florence, Italy. Association for Computational Linguistics.

\bibitem[{Marasini et~al.(2016)Marasini, Quatto, and Ripamonti}]{Marasini2016AssessingTI}
D.~Marasini, P.~Quatto, and E.~Ripamonti. 2016.
\newblock Assessing the inter-rater agreement for ordinal data through weighted indexes.
\newblock \emph{Statistical Methods in Medical Research}, 25:2611 -- 2633.

\bibitem[{Mitchell et~al.(2022)Mitchell, Noh, Li, Armstrong, Agarwal, Liu, Finn, and Manning}]{mitchell-etal-2022-enhancing}
Eric Mitchell, Joseph Noh, Siyan Li, Will Armstrong, Ananth Agarwal, Patrick Liu, Chelsea Finn, and Christopher Manning. 2022.
\newblock \href {https://doi.org/10.18653/v1/2022.emnlp-main.115} {Enhancing self-consistency and performance of pre-trained language models through natural language inference}.
\newblock In \emph{Proceedings of the 2022 Conference on Empirical Methods in Natural Language Processing}, pages 1754--1768, Abu Dhabi, United Arab Emirates. Association for Computational Linguistics.

\bibitem[{Nguyen et~al.(2017)Nguyen, Nguyen, Chu, Thater, and Pinkal}]{nguyen-etal-2017-sequence}
Dai~Quoc Nguyen, Dat~Quoc Nguyen, Cuong~Xuan Chu, Stefan Thater, and Manfred Pinkal. 2017.
\newblock \href {https://aclanthology.org/I17-2007} {Sequence to sequence learning for event prediction}.
\newblock In \emph{Proceedings of the Eighth International Joint Conference on Natural Language Processing (Volume 2: Short Papers)}, pages 37--42, Taipei, Taiwan. Asian Federation of Natural Language Processing.

\bibitem[{Pan et~al.(2020)Pan, Chen, Wu, Liu, Ngo, Kan, Jiang, and Chua}]{pan-etal-2022-multimodal}
Liang-Ming Pan, Jingjing Chen, Jianlong Wu, Shaoteng Liu, Chong-Wah Ngo, Min-Yen Kan, Yugang Jiang, and Tat-Seng Chua. 2020.
\newblock \href {https://doi.org/10.1145/3394171.3413765} {Multi-modal cooking workflow construction for food recipes}.
\newblock In \emph{Proceedings of the 28th ACM International Conference on Multimedia}, MM '20, page 1132–1141, New York, NY, USA. Association for Computing Machinery.

\bibitem[{Pareti et~al.(2014)Pareti, Testu, Ichise, Klein, and Barker}]{pareti2014integrating}
Paolo Pareti, Benoit Testu, Ryutaro Ichise, Ewan Klein, and Adam Barker. 2014.
\newblock Integrating know-how into the linked data cloud.
\newblock In \emph{International Conference on Knowledge Engineering and Knowledge Management}, pages 385--396. Springer.

\bibitem[{Rajani et~al.(2019)Rajani, McCann, Xiong, and Socher}]{rajani-etal-2019-explain}
Nazneen~Fatema Rajani, Bryan McCann, Caiming Xiong, and Richard Socher. 2019.
\newblock \href {https://doi.org/10.18653/v1/P19-1487} {Explain yourself! leveraging language models for commonsense reasoning}.
\newblock In \emph{Proceedings of the 57th Annual Meeting of the Association for Computational Linguistics}, pages 4932--4942, Florence, Italy. Association for Computational Linguistics.

\bibitem[{Robertson and Zaragoza(2009)}]{bm25}
Stephen~E. Robertson and Hugo Zaragoza. 2009.
\newblock \href {https://api.semanticscholar.org/CorpusID:207178704} {The probabilistic relevance framework: Bm25 and beyond}.
\newblock \emph{Found. Trends Inf. Retr.}, 3:333--389.

\bibitem[{Schank and Abelson(1977)}]{schank-abelson-1977-scripts}
R.C. Schank and R.P. Abelson. 1977.
\newblock \href {https://doi.org/10.4324/9780203781036} {Scripts, plans, goals, and understanding: An inquiry into human knowledge structures}.

\bibitem[{Shridhar et~al.(2021)Shridhar, Yuan, C\^ot\'e, Bisk, Trischler, and Hausknecht}]{ALFWorld20}
Mohit Shridhar, Xingdi Yuan, Marc-Alexandre C\^ot\'e, Yonatan Bisk, Adam Trischler, and Matthew Hausknecht. 2021.
\newblock \href {https://arxiv.org/abs/2010.03768} {{ALFWorld: Aligning Text and Embodied Environments for Interactive Learning}}.
\newblock In \emph{Proceedings of the International Conference on Learning Representations (ICLR)}.

\bibitem[{Slaney and Thiébaux(2001)}]{slaney-blocksworld}
John Slaney and Sylvie Thiébaux. 2001.
\newblock \href {https://doi.org/10.1016/S0004-3702(00)00079-5} {Blocks world revisited}.
\newblock \emph{Artificial Intelligence}, 125(1):119--153.

\bibitem[{Tandon et~al.(2020)Tandon, Sakaguchi, Dalvi, Rajagopal, Clark, Guerquin, Richardson, and Hovy}]{tandon-etal-2020-dataset}
Niket Tandon, Keisuke Sakaguchi, Bhavana Dalvi, Dheeraj Rajagopal, Peter Clark, Michal Guerquin, Kyle Richardson, and Eduard Hovy. 2020.
\newblock \href {https://doi.org/10.18653/v1/2020.emnlp-main.520} {A dataset for tracking entities in open domain procedural text}.
\newblock In \emph{Proceedings of the 2020 Conference on Empirical Methods in Natural Language Processing (EMNLP)}, pages 6408--6417, Online. Association for Computational Linguistics.

\bibitem[{Valmeekam et~al.(2023)Valmeekam, Marquez, Olmo, Sreedharan, and Kambhampati}]{valmeekam-etal-2023-planbench}
Karthik Valmeekam, Matthew Marquez, Alberto Olmo, Sarath Sreedharan, and Subbarao Kambhampati. 2023.
\newblock \href {https://arxiv.org/abs/2206.10498} {Planbench: An extensible benchmark for evaluating large language models on planning and reasoning about change}.
\newblock \emph{Preprint}, arXiv:2206.10498.

\bibitem[{Verma et~al.(2023)Verma, Lal, Sinha, Van~Durme, and Poliak}]{verma-etal-2023-evaluating}
Dhruv Verma, Yash~Kumar Lal, Shreyashee Sinha, Benjamin Van~Durme, and Adam Poliak. 2023.
\newblock \href {https://doi.org/10.18653/v1/2023.acl-short.76} {Evaluating paraphrastic robustness in textual entailment models}.
\newblock In \emph{Proceedings of the 61st Annual Meeting of the Association for Computational Linguistics (Volume 2: Short Papers)}, pages 880--892, Toronto, Canada. Association for Computational Linguistics.

\bibitem[{Wei et~al.(2022{\natexlab{a}})Wei, Tay, Bommasani, Raffel, Zoph, Borgeaud, Yogatama, Bosma, Zhou, Metzler, Chi, Hashimoto, Vinyals, Liang, Dean, and Fedus}]{incontext}
Jason Wei, Yi~Tay, Rishi Bommasani, Colin Raffel, Barret Zoph, Sebastian Borgeaud, Dani Yogatama, Maarten Bosma, Denny Zhou, Donald Metzler, Ed~H. Chi, Tatsunori Hashimoto, Oriol Vinyals, Percy Liang, Jeff Dean, and William Fedus. 2022{\natexlab{a}}.
\newblock \href {https://openreview.net/forum?id=yzkSU5zdwD} {Emergent abilities of large language models}.
\newblock \emph{Transactions on Machine Learning Research}.
\newblock Survey Certification.

\bibitem[{Wei et~al.(2022{\natexlab{b}})Wei, Wang, Schuurmans, Bosma, brian ichter, Xia, Chi, Le, and Zhou}]{wei-etal-2022-chain}
Jason Wei, Xuezhi Wang, Dale Schuurmans, Maarten Bosma, brian ichter, Fei Xia, Ed~H. Chi, Quoc~V Le, and Denny Zhou. 2022{\natexlab{b}}.
\newblock \href {https://openreview.net/forum?id=_VjQlMeSB_J} {Chain of thought prompting elicits reasoning in large language models}.
\newblock In \emph{Advances in Neural Information Processing Systems}.

\bibitem[{West et~al.(2024)West, Lu, Dziri, Brahman, Li, Hwang, Jiang, Fisher, Ravichander, Chandu, Newman, Koh, Ettinger, and Choi}]{west-etal-2024-the}
Peter West, Ximing Lu, Nouha Dziri, Faeze Brahman, Linjie Li, Jena~D. Hwang, Liwei Jiang, Jillian Fisher, Abhilasha Ravichander, Khyathi Chandu, Benjamin Newman, Pang~Wei Koh, Allyson Ettinger, and Yejin Choi. 2024.
\newblock \href {https://openreview.net/forum?id=CF8H8MS5P8} {The generative {AI} paradox: {\textquotedblleft}what it can create, it may not understand{\textquotedblright}}.
\newblock In \emph{The Twelfth International Conference on Learning Representations}.

\bibitem[{Wolf et~al.(2020)Wolf, Debut, Sanh, Chaumond, Delangue, Moi, Cistac, Rault, Louf, Funtowicz, Davison, Shleifer, von Platen, Ma, Jernite, Plu, Xu, Le~Scao, Gugger, Drame, Lhoest, and Rush}]{wolf-etal-2020-transformers}
Thomas Wolf, Lysandre Debut, Victor Sanh, Julien Chaumond, Clement Delangue, Anthony Moi, Pierric Cistac, Tim Rault, Remi Louf, Morgan Funtowicz, Joe Davison, Sam Shleifer, Patrick von Platen, Clara Ma, Yacine Jernite, Julien Plu, Canwen Xu, Teven Le~Scao, Sylvain Gugger, Mariama Drame, Quentin Lhoest, and Alexander Rush. 2020.
\newblock \href {https://doi.org/10.18653/v1/2020.emnlp-demos.6} {Transformers: State-of-the-art natural language processing}.
\newblock In \emph{Proceedings of the 2020 Conference on Empirical Methods in Natural Language Processing: System Demonstrations}, pages 38--45, Online. Association for Computational Linguistics.

\bibitem[{Wu et~al.(2024)Wu, Spangher, Alipoormolabashi, Freedman, Weischedel, and Peng}]{wu-etal-2024-understanding}
Te-Lin Wu, Alex Spangher, Pegah Alipoormolabashi, Marjorie Freedman, Ralph Weischedel, and Nanyun Peng. 2024.
\newblock \href {https://arxiv.org/abs/2110.08486} {Understanding multimodal procedural knowledge by sequencing multimodal instructional manuals}.
\newblock \emph{Preprint}, arXiv:2110.08486.

\bibitem[{Yamakata et~al.(2020)Yamakata, Mori, and Carroll}]{yamakata-etal-2020-english}
Yoko Yamakata, Shinsuke Mori, and John Carroll. 2020.
\newblock \href {https://aclanthology.org/2020.lrec-1.638} {{E}nglish recipe flow graph corpus}.
\newblock In \emph{Proceedings of the Twelfth Language Resources and Evaluation Conference}, pages 5187--5194, Marseille, France. European Language Resources Association.

\bibitem[{Zellers et~al.(2019)Zellers, Holtzman, Bisk, Farhadi, and Choi}]{zellers-etal-2019-hellaswag}
Rowan Zellers, Ari Holtzman, Yonatan Bisk, Ali Farhadi, and Yejin Choi. 2019.
\newblock \href {https://doi.org/10.18653/v1/P19-1472} {{H}ella{S}wag: Can a machine really finish your sentence?}
\newblock In \emph{Proceedings of the 57th Annual Meeting of the Association for Computational Linguistics}, pages 4791--4800, Florence, Italy. Association for Computational Linguistics.

\bibitem[{Zhang et~al.(2020{\natexlab{a}})Zhang, Chen, Wang, Song, and Roth}]{zhang-etal-2020-analogous}
Hongming Zhang, Muhao Chen, Haoyu Wang, Yangqiu Song, and Dan Roth. 2020{\natexlab{a}}.
\newblock \href {https://doi.org/10.18653/v1/2020.emnlp-main.119} {Analogous process structure induction for sub-event sequence prediction}.
\newblock In \emph{Proceedings of the 2020 Conference on Empirical Methods in Natural Language Processing (EMNLP)}, pages 1541--1550, Online. Association for Computational Linguistics.

\bibitem[{Zhang et~al.(2020{\natexlab{b}})Zhang, Lyu, and Callison-Burch}]{zhang-etal-2020-reasoning}
Li~Zhang, Qing Lyu, and Chris Callison-Burch. 2020{\natexlab{b}}.
\newblock \href {https://doi.org/10.18653/v1/2020.emnlp-main.374} {Reasoning about goals, steps, and temporal ordering with {W}iki{H}ow}.
\newblock In \emph{Proceedings of the 2020 Conference on Empirical Methods in Natural Language Processing (EMNLP)}, pages 4630--4639, Online. Association for Computational Linguistics.

\bibitem[{Zhang et~al.(2024{\natexlab{a}})Zhang, Xu, Kommula, Callison-Burch, and Tandon}]{zhang-etal-2024-openpi2}
Li~Zhang, Hainiu Xu, Abhinav Kommula, Chris Callison-Burch, and Niket Tandon. 2024{\natexlab{a}}.
\newblock \href {https://aclanthology.org/2024.eacl-long.10} {{O}pen{PI}2.0: An improved dataset for entity tracking in texts}.
\newblock In \emph{Proceedings of the 18th Conference of the European Chapter of the Association for Computational Linguistics (Volume 1: Long Papers)}, pages 166--178, St. Julian{'}s, Malta. Association for Computational Linguistics.

\bibitem[{Zhang et~al.(2023)Zhang, Xu, Yang, Zhou, You, Arora, and Callison-Burch}]{zhang-etal-2023-causal}
Li~Zhang, Hainiu Xu, Yue Yang, Shuyan Zhou, Weiqiu You, Manni Arora, and Chris Callison-Burch. 2023.
\newblock \href {https://doi.org/10.18653/v1/2023.findings-eacl.31} {Causal reasoning of entities and events in procedural texts}.
\newblock In \emph{Findings of the Association for Computational Linguistics: EACL 2023}, pages 415--431, Dubrovnik, Croatia. Association for Computational Linguistics.

\bibitem[{Zhang et~al.(2024{\natexlab{b}})Zhang, Zhang, Hou, Wang, Gu, Clark, Callison-Burch, and Tandon}]{zhang2024proc2pddl}
Tianyi Zhang, Li~Zhang, Zhaoyi Hou, Ziyu Wang, Yuling Gu, Peter Clark, Chris Callison-Burch, and Niket Tandon. 2024{\natexlab{b}}.
\newblock \href {https://arxiv.org/abs/2403.00092} {Proc2pddl: Open-domain planning representations from texts}.
\newblock \emph{Preprint}, arXiv:2403.00092.

\bibitem[{Zheng et~al.(2024)Zheng, Mishra, Zhang, Chen, Chen, Nova, Hou, Cheng, Le, Chi, and Zhou}]{zheng-etal-2024-naturalplan}
Huaixiu~Steven Zheng, Swaroop Mishra, Hugh Zhang, Xinyun Chen, Minmin Chen, Azade Nova, Le~Hou, Heng-Tze Cheng, Quoc~V. Le, Ed~H. Chi, and Denny Zhou. 2024.
\newblock \href {https://arxiv.org/abs/2406.04520} {Natural plan: Benchmarking llms on natural language planning}.
\newblock \emph{Preprint}, arXiv:2406.04520.

\end{thebibliography}
